%% file: iclr2027_conference.tex
\documentclass{article} 
\usepackage{iclr2027_conference,times}

\input{math_commands.tex}

\usepackage{float}
\usepackage{hyperref}
\usepackage{url}
\usepackage{xspace}
\usepackage{booktabs}
\usepackage{graphicx}
\usepackage{xcolor}
\usepackage[most]{tcolorbox}
\usepackage{caption}
\definecolor{IdeaRewardPromptBlue}{HTML}{5BB5E6}
\definecolor{IdeaRewardPromptBackground}{HTML}{EEF8FD}
\tcbset{idearewardprompt/.style={
  enhanced,
  breakable,
  colback=IdeaRewardPromptBackground,
  colframe=IdeaRewardPromptBlue,
  boxrule=0.7pt,
  arc=1.5mm,
  left=2.5mm,
  right=2.5mm,
  top=2mm,
  bottom=2mm,
  listing only,
  listing options={
    basicstyle=\ttfamily\footnotesize,
    breaklines=true,
    breakautoindent=false,
    breakindent=0pt,
    columns=fullflexible,
    keepspaces=true
  }
}}

\newcommand{\IdeaReward}{\textsc{IdeaReward}\xspace}
\newcommand{\IdeaRewardBench}{\textsc{IdeaRewardBench}\xspace}

\title{IdeaReward: Benchmarking and Training \\Reward Models for Research Idea Generation}

\author{Amy Xin\thanks{Equal contribution.}, Xinming Xu\footnotemark[1], Yunjia Qi, Bin Xu, Lei Hou, Juanzi Li \\
Tsinghua University \\
\texttt{\{xin-x25, xu-zh23\}@mails.tsinghua.edu.cn}
}

\iclrfinalcopy 
\begin{document}

\maketitle
\lhead{}

\begin{abstract}
  Research idea generation aims to produce research ideas that can be developed into publication-worthy research projects.
  However, as large language model (LLM)-based systems generate research ideas at scale, the central bottleneck shifts from producing candidate ideas to reliably evaluating, selecting, and improving them before costly research execution.
  Reward models offer a promising solution for this goal by learning to assign quality-aware reward scores from preference data.
  In this paper, we present a systematic study of reward modeling for research idea generation. 
  We present \IdeaRewardBench, the first multi-signal, living benchmark for research idea evaluation.
  \IdeaRewardBench contains 3,056 idea preference instances grounded in complementary signals of idea quality, 
  and is updated semiannually to reflect the latest research frontier. 
  We evaluate a comprehensive set of 23 baselines on the full \IdeaRewardBench, 
  and find that all baselines struggle to distinguish high-quality research ideas: the best baseline achieves only 62.8\% pairwise accuracy on the review-grounded split and 60.6\% on the impact-grounded split, where 50\% is the random baseline.
  We then present \IdeaReward, the first reward model for research idea generation, trained on 19,670 high-quality idea preference pairs constructed by a fully automated pipeline.
  \IdeaReward establishes state-of-the-art performance on \IdeaRewardBench,
  outperforming baselines based on much larger LLMs.
  We further show that \IdeaReward improves downstream test-time best-of-N idea selection.
  We release our data, models, and code to facilitate future research \footnote{\url{https://github.com/THU-KEG/IdeaReward}}.
\end{abstract}

\section{Introduction}

Research idea generation aims to automatically generate research idea proposals that can be developed into publication-worthy projects and contribute meaningfully to the research frontier~\citep{scimon,novelideas}.
Large language model (LLM)-based systems 
have demonstrated promising potential in research idea generation \citep{researchagent,novelideas,aiscientist,nova,chainofideas,aiscientistv2,airesearcher,coscientist,aris}.
Nevertheless, 
\textit{reward modeling} for research idea generation remains largely underexplored.
Ideally, given a candidate pool of research ideas, reward models should accurately evaluate each candidate's quality, distinguish strong ideas from weak ones,
 and provide feedback signals 
at both training and inference time.
However, existing reward models primarily capture general response qualities rather than research idea quality \citep{rewardmodelsurvey,rewardbench}, while LLM-based evaluators often provide unreliable rewards by struggling to judge idea novelty, feasibility, and significance \citep{novbench,scientificnoveltylimits,llmjudge}.
As AI systems begin to generate research ideas 
at scale,
the 
bottleneck shifts from producing candidate ideas to learning reliable reward signals that can evaluate, select, and improve those ideas before costly execution.
There is therefore an urgent need to study reward modeling for research idea generation.

A systematic study of reward modeling for research idea generation faces two key challenges.
\textbf{(1) Building a timely, accurate, and unified benchmark for idea evaluation.}
Studying idea reward models first requires a benchmark that can reliably measure how well they distinguish higher-quality ideas from lower-quality ones.
However, evaluation of idea quality depends on an evolving research frontier,
requires complementary signals that collectively reflect idea quality, 
and demands a unified protocol for fairly comparing heterogeneous idea evaluators, including scalar reward models and LLM-based judges.
\textbf{(2) Collecting large-scale, high-quality training data for idea reward models.}
Training idea reward models requires large quantities of idea-level preference data, yet open-source research idea proposals are scarce and costly to annotate.
Moreover, idea quality is inherently multidimensional, spanning dimensions such as novelty, feasibility, and significance.
Training data must therefore be scalable, high quality, and dimension-aware.

In this paper, we present the first systematic study of reward modeling for research idea generation through both a benchmark and a trained reward model.
We first present \IdeaRewardBench, a multi-signal, living benchmark for research idea evaluation.
Overall, \IdeaRewardBench comprises 3,056 high-quality idea preference instances in two formats: pairwise idea preferences, each with one chosen and one rejected research idea, and quadruple idea preferences, each with a strict preference order over four research ideas.
\IdeaRewardBench contains three main subsplits, 
each defined by a distinct source of quality signal:
(i) a review-grounded split with 1,865 instances, built from submissions to the latest ICLR, ICML, and NeurIPS conferences, where preferences are derived from open peer reviews;
(ii) an impact-grounded split with 1,091 instances, built from arXiv papers released 
within the past year and a half, where preferences are derived from comparative citation data; and
(iii) an annotation-grounded split with 100 instances, built from human and LLM-generated research ideas under a controlled setting, where preferences are derived from direct expert annotation.
We also provide an additional, special ``novelty attack'' split to test whether idea evaluators can recognize already published ideas and penalize them for lacking novelty.
We will update \IdeaRewardBench continuously on a semiannual schedule with latest data to keep it aligned with the evolving research frontier.
By integrating preferences grounded in complementary quality signals from  conference peer review, citation impact, and expert annotation, \IdeaRewardBench captures the multifaceted nature of idea quality and reduces bias from any single source.
Finally, by conducting evaluation based on the predicted idea preference order, \IdeaRewardBench supports evaluation of any reward model or LLM-based evaluator that outputs either a scalar score for a single idea or a rank of idea preferences.

We also present \IdeaReward, a reward model for research idea generation, trained on 19,670 automatically constructed idea preference pairs.
Mirroring the complementary signals in \IdeaRewardBench, we build \IdeaReward's training data from three sources:
(i) review-based data, 
derived from historical conference submissions and open peer reviews 
preceding the benchmark period;
(ii) impact-based data,
 derived from historical arXiv papers preceding the benchmark period 
 using comparative citation data; and
(iii) controlled synthetic data,
 generated through automated preference construction methods including strong-weak LLM pairing, LLM-judge-based preference assignment, and dimension-specific degradation.
Together, these sources enable fully automated and scalable training data construction, while dimension-specific degradation helps \IdeaReward learn dimension-aware judgments of idea quality.

We evaluate \IdeaReward against 23 baselines on the full \IdeaRewardBench, including general reward models, naive LLM judges, and specialized idea evaluators.
We find that all baselines struggle to distinguish high-quality ideas on \IdeaRewardBench, including specialized idea evaluators that use complex agentic workflows or trained on idea-related data. The best baseline achieves only 62.8\% pairwise accuracy on the review-grounded split and 60.6\% on the impact-grounded split, where 50\% corresponds to random guessing.
Finally, we show that \IdeaReward achieves the best performance across all eight metrics among these 23 baselines, and further validate its effectiveness in a downstream test-time best-of-N idea selection application.


Our contributions are threefold:

(1) We present \IdeaRewardBench, the first living benchmark for research idea evaluation.
\IdeaRewardBench comprises 3,056 idea preference instances grounded in multiple complementary signals of idea quality.
Evaluating 23 baselines on the full \IdeaRewardBench reveals substantial limitations in existing idea evaluators, highlighting the benchmark's difficulty and diagnostic value.

(2) We present \IdeaReward, the first reward model for research idea generation.
\IdeaReward is trained with a fully automated and scalable data construction pipeline, and achieves state-of-the-art performance across all eight metrics among the 23 full-benchmark baselines.

(3) We conduct extensive experiments and analyses on \IdeaReward and evaluated baselines, revealing key findings including that existing evaluators struggle to distinguish borderline from stronger ideas, models trained on a single quality signal do not yield robust and transferable idea judgment, and retrieval-equipped evaluators consistently misjudge already published ideas as novel.


\section{IdeaRewardBench: A Multi-signal, Living Benchmark for Research Idea Evaluation}
In this section, we present the construction of \IdeaRewardBench.
We first define task formulation (\S~\ref{sec:2_task_formulation}), then present data construction methods (\S~\ref{sec:2_benchmark_data_construction}) and metric design (\S~\ref{sec:2_metric_design}).

\begin{figure}[b]
\centering
\begin{tcolorbox}[
  width=0.98\linewidth,
  colback=gray!3,
  colframe=gray!60!black,
  colbacktitle=gray!65!black,
  coltitle=white,
  title={Research Idea Template},
  fonttitle=\bfseries\footnotesize,
  fontupper=\footnotesize,
  boxrule=0.8pt,
  arc=2pt,
  left=6pt,
  right=6pt,
  top=5pt,
  bottom=5pt,
  toptitle=3pt,
  bottomtitle=3pt
]
\textbf{1. Title:} A concise research-paper-style title.

\textbf{2. Research Question:} A concrete research question that motivates the proposed idea.

\textbf{3. Motivations:} A detailed explanation of why the research question matters and what gaps it addresses.

\textbf{4. Methods:} A detailed description of the proposed approach.

\textbf{5. Experiment Plan:} A concrete plan for evaluating the proposed approach, including concrete tasks or datasets, metrics, baselines, ablations, and other analysis experiments.
\end{tcolorbox}
\caption{The five-field template used to represent a research idea.}
\label{fig:idea_template}
\vspace{-0.8em}
\end{figure}

\subsection{Task Formulation}
\label{sec:2_task_formulation}

A research idea is a minimal, concise research proposal that should support execution into a complete research project and ideally a high-quality publication.
Following prior work
 ~\citep{aiscientist,researchagent,novelideas,innoeval}, we represent an idea \(I\) as:
\begin{equation}
\label{eq:idea_schema}
I = \langle t, q, m, a, e \rangle,
\end{equation}
where \(t\) denotes the research title, \(q\) denotes the target research question, \(m\) denotes the research motivations, \(a\) denotes the proposed methodological approach, and \(e\) denotes the experiment plan.
The idea template is expanded in Figure~\ref{fig:idea_template}.
Research idea evaluation aims to distinguish research ideas of different quality.
Given a set of candidate ideas \(\mathcal{I}=\{I_1,\ldots,I_n\}\), an idea evaluator \(E\) induces a predicted preference order \(\hat{\pi}_E\) over \(\mathcal{I}\), either from ranked pointwise scores or direct comparative judgments.
The objective is to make \(\hat{\pi}_E\) consistent with the gold quality order \(\pi_Q\) assigned by the idea quality function \(Q(I)\).
\IdeaRewardBench contains two types of idea preference instances:

\textbf{Idea Preference Pair.}
An idea preference pair consists of a chosen idea \(I^{+}\) and a rejected idea \(I^{-}\), where the chosen idea has higher overall quality:
\begin{equation}
\label{eq:idea_pair}
Q(I^{+}) > Q(I^{-}).
\end{equation}
The objective is to predict \(I^{+}\) as preferred over \(I^{-}\).

\textbf{Idea Preference Quadruple.}
An idea preference quadruple consists of four candidate ideas with a strict quality order:
\begin{equation}
\label{eq:idea_quadruple}
Q(I^{(1)}) > Q(I^{(2)}) > Q(I^{(3)}) > Q(I^{(4)}).
\end{equation}
The objective is to identify the highest-quality idea \(I^{(1)}\) and to recover the relative preference order among all four ideas.


\subsection{Benchmark Data Construction}
\label{sec:2_benchmark_data_construction}

As research idea evaluation is inherently forward-looking and multifaceted, relying on any single idea quality signal can introduce systematic bias.
\IdeaRewardBench 
triangulates idea quality through three complementary quality signals that differ in both temporal horizon and evaluation perspective:
\textbf{(1) review-grounded data}, which captures idea quality through peer reviews at paper submission time;
\textbf{(2) impact-grounded data}, which captures idea quality through 
community impact over a longer horizon; and
\textbf{(3) annotation-grounded data}, which captures idea quality through 
expert judgments at idea formulation time.
To keep \IdeaRewardBench aligned with the evolving research frontier, we update its data on a semiannual schedule.
Together, these designs make \IdeaRewardBench a living, multi-signal benchmark for comprehensive research idea evaluation.
Because the review and impact-grounded splits use paper-level
outcomes as proxies for idea quality, we conduct a pilot validation on 200 instances sampled from the benchmark construction pool and find 84.5\% top-choice agreement between paper-level and human idea-level quality preferences, as detailed in Appendix~\ref{sec:appendix_preliminary_experiment}.

\textbf{Review-grounded Data.}
\label{sec:2_review_grounded_data}
The review-grounded split constructs idea preferences from conference paper submissions and peer review data.
We construct idea preference instances from submissions to the latest NeurIPS, ICLR, and ICML conferences, namely NeurIPS 2025, ICLR 2026, and ICML 2026.
For each paper, we collect its full manuscript, final decision, and peer reviews from OpenReview, then use an automated LLM-based pipeline to extract and anonymize a structured idea card following the template in Figure~\ref{fig:idea_template}.
Details of the LLM-based idea extraction pipeline are provided in Appendix~\ref{sec:appendix_llm_idea_extraction}.
To ensure comparability, we group ideas within the same conference and primary research area.
Each idea preference pair consists of one accepted submission and one rejected submission sampled from the same group.
To reduce noise from borderline cases, we compute the gap between the average human reviewer 
scores of the accepted and rejected submissions, and retain only pairs with a score gap of at least \(0.5\).
Similarly, each idea preference quadruple is constructed by sampling one idea from each of four quality pools: strong accept, standard accept, standard reject, and strong reject.
We define strong accept as oral or spotlight papers, standard accept as poster papers, standard reject as rejected papers whose average review score is equal or above the conference borderline-reject threshold, and strong reject as rejected papers below that threshold.
We obtain 1,580 pair instances and 285 quadruple instances, with a total of 1,865 review-grounded instances.

\begin{figure*}[t]
\centering
\includegraphics[width=\textwidth]{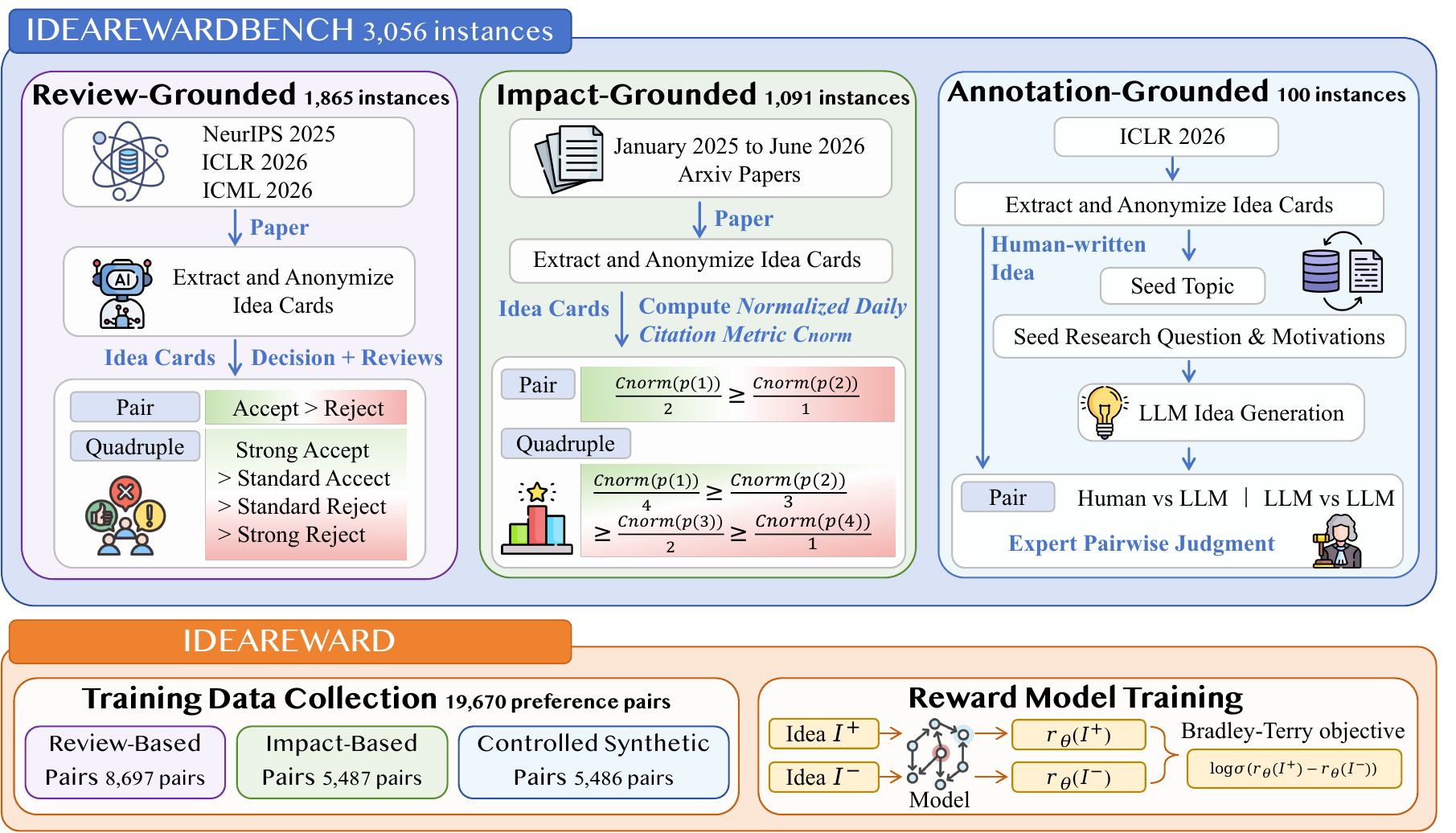}
\caption{Overview of \IdeaRewardBench construction and \IdeaReward training.}
\label{fig:ideareward_overview}
\end{figure*}

\textbf{Impact-grounded Data.}
\label{sec:2_impact_grounded_data}
Conference peer reviews reflect expert judgment at paper submission time, while many research ideas reveal their value only through later community impact.
We therefore construct impact-grounded idea preference instances from paper citation data.
We collect papers from the AI, CL, and LG categories of arXiv from the past year and a half, namely January 2025 to July 2026, excluding technical reports and papers already featured in the review-grounded split.
We group papers within the same arXiv category under three criteria: they are first released within the same three-month window, do not cite each other, and have SPECTER2 embedding similarity above \(0.95\).
For each paper \(p\) in a comparison group \(\mathcal{G}\), we compute a normalized daily citation metric:
\begin{equation}
\label{eq:normalized_daily_citation}
C_{\mathrm{norm}}(p;\mathcal{G})
=
\frac{C(p)}{D(p)} \cdot \max_{p' \in \mathcal{G}} D(p'),
\end{equation}
where \(C(p)\) is the current citation count of paper \(p\), \(D(p)\) is the number of days since \(p\) was first released, and \(\max_{p' \in \mathcal{G}} D(p')\) is the exposure time of the earliest released paper in the group.
For each paper, we then perform the same idea-card extraction and anonymization pipeline as the review-grounded split.
Idea preference pairs are constructed by sampling a chosen paper \(p^+\) and a rejected paper \(p^-\) from the same group \(\mathcal{G}\), requiring the ratio of \(C_{\mathrm{norm}}(p^+; \mathcal{G})\) to \(C_{\mathrm{norm}}(p^-;\mathcal{G})\) to be at least \(2:1\), and \(p^+\) to have at least \(30\) raw citations.
Idea preference quadruples are constructed by sampling four papers \(p^{(1)},\ldots,p^{(4)}\) from the same group \(\mathcal{G}\), requiring the ratio of \(C_{\mathrm{norm}}\) from \(p^{(1)}\) to \(p^{(4)}\) to be at least \(4:3:2:1\), and \(p^{(1)}\) to have at least \(30\) raw citations.
We obtain \(827\) pair instances and \(264\) quadruple instances, yielding \(1{,}091\) impact-grounded instances in total.

\textbf{Annotation-grounded Data.}
\label{sec:2_annotation_grounded_data}
The annotation-grounded split evaluates idea quality at proposal time, without relying on an executed paper.
We design this split via controlled human-vs-LLM and LLM-vs-LLM comparisons.
\textbf{(1) Human-vs-LLM Subsplit.}
We first sample \(50\) human-written ideas from unused ICLR 2026 submissions.
Then, we use half of the human ideas to derive seed topics for topic-guided LLM idea generation and the other half to extract seed research questions and motivations for LLM idea completion, yielding \(50\) human-vs-LLM pairs.
\textbf{(2) LLM-vs-LLM Subsplit.}
We use an additional \(25\) seed research questions and motivations extracted from unused ICLR 2026 submissions, together with \(5\) manually designed seed topics. 
We manage a pool of six candidate LLMs covering diverse large commercial models, large open-source models, and smaller open-source models.
For research-question-seeded comparisons, we randomly select two LLMs to independently complete the idea.
For topic-seeded comparisons, each LLM generates five candidate ideas per topic, which are then randomly paired.
This yields \(50\) LLM-vs-LLM pairs.
Finally, two human experts independently select the better idea in each pair and resolve disagreements through discussion.
We obtain 100 annotation-grounded idea preference pair instances in total.


\subsection{Metric Design}
\label{sec:2_metric_design}

We evaluate performance on \IdeaRewardBench using three tie-aware metrics.

\paragraph{Pair Accuracy.}
For each idea preference pair \((I^+, I^-)\), where \(I^+\) is the gold preferred idea and \(u(I)\) denotes the evaluator-predicted idea quality, pair accuracy is:
\begin{equation}
\label{eq:pair_accuracy}
\mathrm{PairAcc}(I^+, I^-)=
\begin{cases}
1, & u(I^+) > u(I^-), \\
0.5, & u(I^+) = u(I^-), \\
0, & u(I^+) < u(I^-).
\end{cases}
\end{equation}
The final score is averaged over all pair instances, with a random baseline of 0.50.

\paragraph{Quadruple Top-1 Accuracy.}
For each idea preference quadruple \(Q(I^{(1)}) > Q(I^{(2)}) > Q(I^{(3)}) > Q(I^{(4)})\), top-1 accuracy measures whether the evaluator assigns the highest predicted quality to \(I^{(1)}\).
If \(I^{(1)}\) is tied among \(k\) top-scored ideas, the instance receives \(1/k\); otherwise it receives \(0\).
The final score is averaged over all quadruple instances, with a random baseline of 0.25.

\paragraph{Quadruple Pairwise Ranking Accuracy.}
Pairwise ranking accuracy (PRA) measures consistency with all six gold pairwise relations in a quadruple.
For each pair \((I^{(a)}, I^{(b)})\) with \(a<b\), let \(s_{ab}=1\) if \(u(I^{(a)})>u(I^{(b)})\), \(0.5\) if tied, and \(0\) otherwise.
Then, the quadruple-level PRA is
\begin{equation}
\label{eq:pra}
\mathrm{PRA} = \frac{1}{6} \sum_{1 \leq a < b \leq 4} s_{ab}.
\end{equation}
The final score is averaged over all quadruple instances.

\section{IdeaReward: A Reward Model for Research Idea Generation}

In this section, we present the training of \IdeaReward.
We first describe its training data construction (\S~\ref{sec:3_training_data_construction}), then present details on model training (\S~\ref{sec:3_training_ideareward}).

\subsection{Training Data Construction}
\label{sec:3_training_data_construction}
We train \IdeaReward on 19,670 automatically constructed idea preference pairs.
Mirroring \IdeaRewardBench, training data is constructed from three sources: \textbf{(1) review-based pairs}, derived from historical conference submissions and OpenReview peer reviews;
\textbf{(2) impact-based pairs}, derived from historical arXiv papers using comparative citation data; and \textbf{(3) controlled synthetic pairs}, generated through topic-aware human-vs-LLM pairing, strong-vs-weak LLM pairing, LLM-judge preference assignment, and dimension-specific quality control.
 Paper-derived training and benchmark data use strictly disjoint source papers to prevent data leakage.
More details on training data construction are described in Appendix~\ref{sec:appendix_training_data_construction}.

\subsection{Training \IdeaReward}
\label{sec:3_training_ideareward}
\IdeaReward parameterizes a scalar reward function \(r_\theta(I)\) over a research idea \(I\).
Given a training set \(\mathcal{D}_{\mathrm{train}}\) of preference pairs \((I^+, I^-)\), where \(I^+\) is the preferred idea, we train \IdeaReward with the Bradley--Terry objective:
\begin{equation}
\label{eq:ideareward_bt}
\mathcal{L}_{\mathrm{BT}}(\theta)
=
-\frac{1}{|\mathcal{D}_{\mathrm{train}}|}
\sum_{(I^+, I^-) \in \mathcal{D}_{\mathrm{train}}}
\log \sigma\!\left(r_\theta(I^+) - r_\theta(I^-)\right).
\end{equation}
At inference time, \IdeaReward scores each idea independently with \(r_\theta(I)\), and idea preferences are obtained by comparing these scalar rewards.
We instantiate \IdeaReward with both Llama-3.1-8B-Instruct and Qwen3-8B backbones, and train each model for one epoch with a global batch size of \(256\) and a maximum sequence length of \(16{,}384\), using learning rates of \(9\times10^{-6}\) for Llama-3.1-8B-Instruct and \(2\times10^{-5}\) for Qwen3-8B.

\section{Experiments}

\input{tables/main_results}

\subsection{Main Experiments}

\textbf{Experimental Setup.}
We evaluate \IdeaReward-Llama and \IdeaReward-Qwen against 23 baselines on the full \IdeaRewardBench, spanning three method categories.
(1) General reward models, which are trained for general LLM alignment.
 We evaluate pointwise reward models ArmoRM-Llama3-8B~\citep{armorm}, QRM-Llama3.1-8B-v2~\citep{qrm}, Skywork-Reward-V2-Llama-3.1-8B~\citep{skyworkrewardv2}, InternLM2-20B-Reward~\citep{internlm2}, and DeepSeek-GRM-16B~\citep{deepseekgrm}, as well as pairwise reward models PairRM~\citep{pairrm}, pair-preference-model-LLaMA3-8B~\citep{rlhflow,slichf}, and RRM-7B~\citep{rrm}.
(2) Naive LLM judges, which uses frozen LLMs as 
idea evaluators using a fixed 
idea review prompt.
We evaluate Llama3.1-8B-Instruct~\citep{llama3}, Qwen3.8-27B~\citep{qwen38}, GLM-5.2~\citep{glm52}, DeepSeek-v4-pro~\citep{deepseekv4}, and GPT-5.5~\citep{openai2026models}.
(3) Specialized idea evaluators, which further include: idea evaluator modules from agentic research systems AI Scientist~\citep{aiscientist}, AutoResearchClaw~\citep{autoresearchclaw}, ResearchAgent~\citep{researchagent}, InternAgent~\citep{internagent}, EvoScientist-skills~\citep{evoscientist}, and ARIS~\citep{aris}; the dedicated idea evaluation system Intern-Atlas~\citep{internatlas}; and specialized paper or abstract review models, including CycleReviewer~\citep{cycleresearcher}, DeepReviewer~\citep{deepreview}, and SciJudge~\citep{scijudge}.
We fix the LLM backbone of specialized idea evaluator baselines that rely on external LLMs to GLM-5.2, and set the search cutoff for retrieval-equipped baselines to the day before the corresponding paper's initial release date to prevent data leakage.
We additionally evaluate the search-intensive InnoEval framework on a fixed random subset; its results and evaluation details are reported in Appendix~\ref{sec:appendix_evaluation_details}.


\begin{figure*}[!t]
    \centering
    \includegraphics[width=1\textwidth]{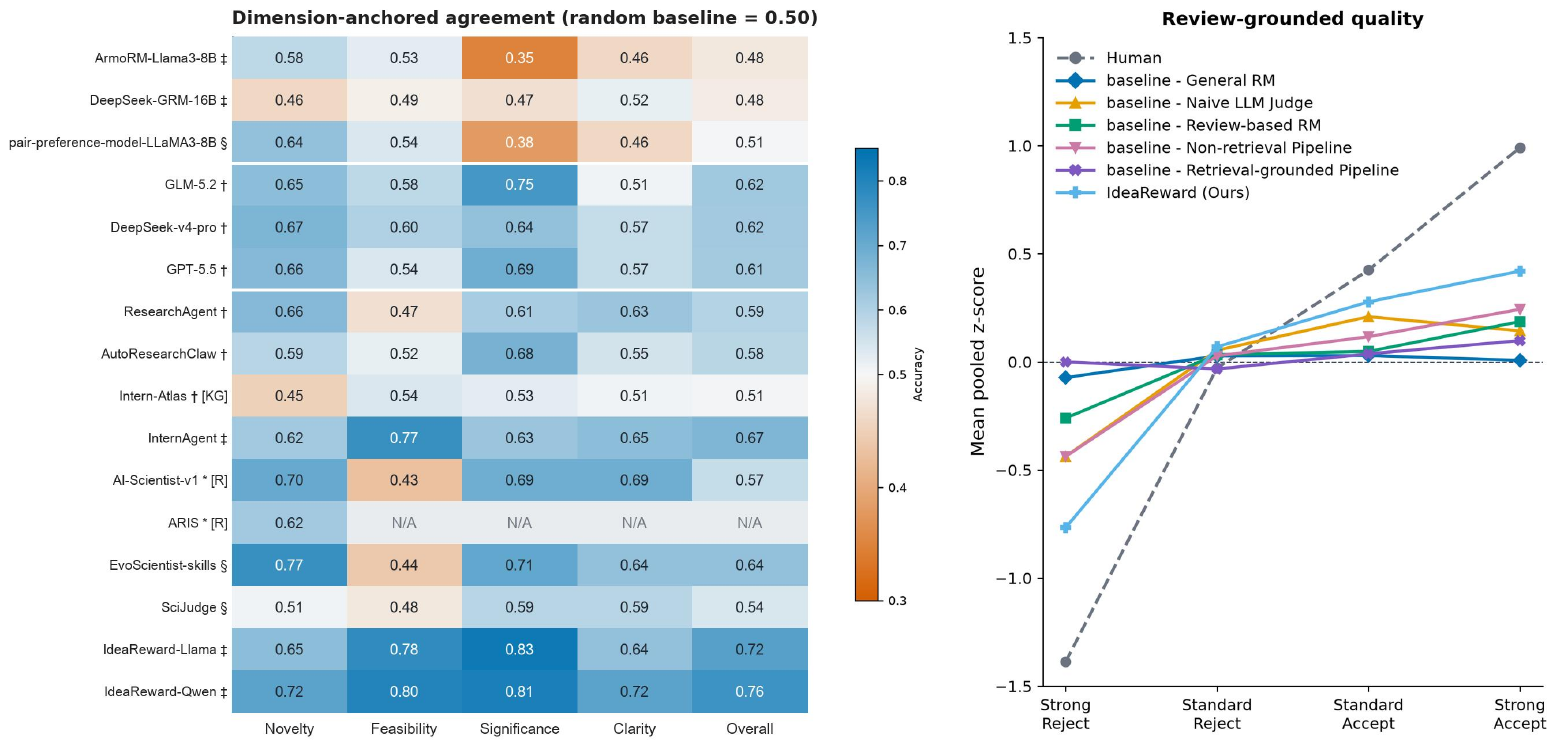}
    \caption{
      \textbf{Dimensional analysis and score distribution analysis results.}
    \textbf{Left:} Dimension-anchored pairwise accuracy.
    \(\dagger\) denotes methods measured by natively supported dimensional scores; \(\ddagger\) denotes methods measured by overall-score; \(\star\) denotes dimensional scores when available and otherwise an overall-score fallback; \(\S\) denotes pairwise methods. [R] indicates equipped with web search, and [KG] indicates equipped with knowledge-graph retrieval.
    \textbf{Right:} Score distribution analysis results for review-grounded quadruples.
    }
    \label{fig:dimension_quality_combined}
\end{figure*}

\textbf{Experimental Results.}
Table~\ref{tab:main_results} reports the main results on \IdeaRewardBench.
Our key observations are as follows:
\textbf{(1) Existing baselines perform poorly on \IdeaRewardBench.} 
The best baseline achieves only \(62.8\%\) pairwise accuracy on the review-grounded split and \(60.6\%\) on the impact-grounded split, where \(50\%\) corresponds to random guessing.
This highlights the need for systematic benchmarking to assess and improve idea evaluation.
\textbf{(2) General reward models fail to capture research idea quality.}
Both pointwise and pairwise reward models remain near-random performance on the review and impact-grounded splits.
They also show a strong bias toward LLM-generated content, with accuracies substantially below \(50\%\) on the Human-vs-LLM subsplit.
This indicates that research idea evaluation requires dedicated reward modeling rather than general preference modeling.
\textbf{(3) Existing training-based idea evaluators do not generalize robustly across quality signals.}
CycleReviewer and DeepReviewer perform weakly across all splits;
SciJudge, which is trained on citation-based paper abstract preferences, achieves the highest impact-grounded pair accuracy, but does not transfer this advantage to the other data splits.
This indicates that idea reward modeling should be treated as a multi-signal modeling problem.
\textbf{(4) External knowledge retrieval provides unstable gains.}
Intern-Atlas and ARIS remain near random across most subsplits.
On the other hand, within its evaluated subset, InnoEval achieves strong performance in review-grounded pair accuracy and LLM-vs-LLM accuracy, but fails to generalize its performance to other splits and metrics.
This indicates that supporting external knowledge retrieval alone does not necessarily improve idea evaluation, and must be combined with robust retrieval and reasoning strategies to achieve consistent gains.
\textbf{(5) \IdeaReward achieves state-of-the-art performance across all metrics among the full-benchmark baselines.}
\IdeaReward-Qwen outperforms all 23 baselines on each of the eight reported metrics, including the much larger GPT-5.5-based naive LLM judge and GLM-5.2-based specialized idea evaluators.
This demonstrates the effectiveness of the multi-signal data construction and reward-model training recipe of \IdeaReward.
Additional analyses in Appendix~\ref{sec:appendix_shortcut_analysis} further show that \IdeaReward's strong performance is not explained by hacking simple heuristic shortcuts.

\subsection{Analysis}

\paragraph{Score Distribution Analysis: Can Idea Evaluators Separate Research Ideas as Humans Do?}
\label{sec:score_distribution}


As Table~\ref{tab:main_results} reports only preference-level performance, we further examine how well \IdeaReward and the baselines separate ideas across quality levels.
Using the idea-level predicted scores of the 285 review-grounded quadruples, we perform z-score normalization for each method and conference as \(z_{m,c,i}=(s_{m,c,i}-\mu_{m,c})/\sigma_{m,c}\).
We group baselines by method type and aggregate their normalized scores, as shown in Figure~\ref{fig:dimension_quality_combined} (right).
We conduct the same analysis on the 264 impact-grounded quadruples, normalizing each method's scores within each arXiv primary category, with results shown in Figure~\ref{fig:impact_sensitivity_appendix}.
Across both splits, \IdeaReward exhibits the clearest standardized score separation across idea-quality categories and most closely follows the monotonic trend in human scores.
While most baseline categories clearly separate strong rejects from standard rejects, they struggle to distinguish finer quality differences among higher-quality ideas.

\begin{figure}[t]
    \centering
    \includegraphics[width=1\linewidth]{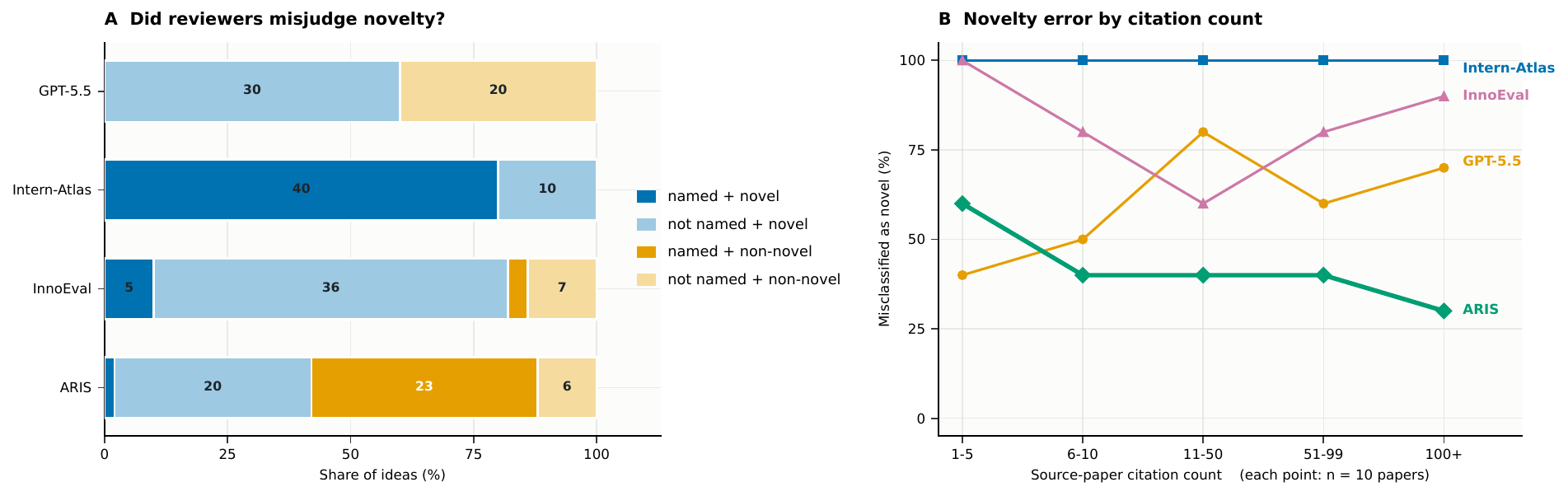}
    \caption{Evaluation results on the novelty attack split.}
    \label{fig:novelty_attack}
\end{figure}

\textbf{Dimensional Analysis: Which Dimensions of Idea Quality Can Idea Evaluators Capture?}
As idea quality is inherently multidimensional, 
we investigate which quality dimensions \IdeaReward and baselines capture.
We map the official review criteria of NeurIPS, ICLR, and ICML into four unified dimensions: novelty, feasibility, significance, and clarity, with more details described in Appendix~\ref{sec:appendix_dimensional_analysis}.
Using collected conference peer reviews, we construct 200 dimension-anchored pairs from the review-grounded split, 50 per dimension, requiring the chosen idea to have a higher average review score on the corresponding dimension.
We evaluate \IdeaReward against a core subset of baselines, using each system's native dimension score when available and its overall score otherwise.
Results are shown in Figure~\ref{fig:dimension_quality_combined} left.
Except InternAgent, all baselines exhibit particularly weak judgment on feasibility, while \IdeaReward performs strongly, indicating that feasibility judgment can be learned.
\IdeaReward and most baselines perform best on significance, likely because high-significance ideas often contain linguistically salient contribution framing.
Interestingly, general reward models perform worst on significance, further highlighting their misalignment with the idea evaluation task.




\textbf{Novelty Attack: Can Idea Evaluators Recognize Previously Published Ideas?}
As more idea generation and evaluation frameworks incorporate external retrieval, we ask whether they can recognize published ideas and correctly penalize them as non-novel.
This capability is essential for scientific idea evaluation: a reliable idea evaluator should retrieve highly similar prior work and reject ideas that have already been explored.
We construct a novelty attack split with 50 anonymized idea cards extracted from arXiv papers released between January and October 2025, balanced across months and five citation-count buckets.
We evaluate two outcomes: whether the method identifies the source paper, and whether it misjudges the idea as novel by assigning a normalized novelty score of at least 0.6.
We evaluate four representative LLM-based methods: GPT-5.5-based naive LLM judge with no external retrieval, Intern-Atlas with knowledge graph-based retrieval, and InnoEval and ARIS with web search-based retrieval.
All papers are released before the GPT-5.5 cutoff date of December 2025 and included in the Intern-Atlas knowledge graph.
We intentionally set the web-search cutoff for InnoEval and ARIS to August 2026, allowing them to retrieve the original papers.

Figure~\ref{fig:novelty_attack} shows that all methods frequently misclassify published ideas as novel, with error rates from 42\% for ARIS to 100\% for Intern-Atlas.
Despite using web search, InnoEval and ARIS identify the original paper in only 14\% and 48\% of cases, respectively.
Intern-Atlas retrieves the original paper for 80\% of ideas, yet still assigns high novelty scores in all cases.
The right panel further shows that higher citation impact reduces novelty misjudgement only for ARIS, while other baselines show no consistent trend.
These results reveal substantial room for improving retrieval-equipped idea evaluators.




\section{Application: Test-time Best-of-N Research Idea Selection}
We evaluate \IdeaReward for test-time best-of-N research idea selection.
We curate 60 expert-crafted seed topics and generate seven candidate ideas per topic, using GLM-5.2 for 45 topics and DeepSeek-v4-pro for 15 topics.
For each topic, \IdeaReward-Qwen, InternLM2-20B-Reward, GPT-5.5, and GLM-5.2 each select one best idea.
Two human experts independently compare \IdeaReward's selection against each baseline's selection and resolve disagreements through discussion.
We treat identical selections as hard ties and expert-judged comparable selections as soft ties.
\IdeaReward achieves non-tie win rates of 81.6\% against InternLM2-20B-Reward, 63.3\% against GPT-5.5, and 68.8\% against GLM-5.2.
The advantage over InternLM2-20B-Reward is statistically significant, the result against GLM-5.2 is at the boundary of statistical significance, and the favorable result against GPT-5.5 is not statistically significant.
Detailed results are reported in Appendix~\ref{sec:appendix_bon_significance}.
This experiment evaluates pre-execution idea selection rather than guaranteeing downstream project success.

\begin{figure}[t]
\centering
\includegraphics[width=0.9\linewidth]{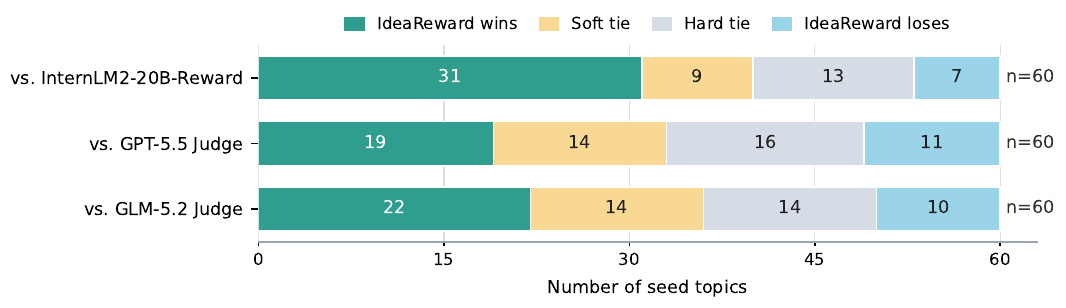}
\caption{Head-to-head results for BoN idea selection over \(60\) seed topics.}
\label{fig:bon_selection}
\end{figure}

\section{Related Work}

\textbf{Research Idea Generation.}
The scientific reasoning capabilities of LLMs have driven increasing interest in automated research idea generation.
Specialized idea generation systems retrieve, recombine, or iteratively refine research ideas~\citep{scimon,novelideas,researchagent,nova,chainofideas},
while end-to-end AI scientist systems extend idea proposal toward experimentation, review, and result presentation~\citep{aiscientist,aiscientistv2,airesearcher,coscientist,cycleresearcher,evoscientist,aris,eurekagent}.
Despite this progress, LLM-generated ideas can still be limited in novelty, significance, feasibility, and diversity~\citep{novelideas,unlockideas,aiscientistv2,scientificnoveltylimits}.
Moreover, current LLM-based self-review and idea judging remain unreliable for assessing scientific novelty and idea quality~\citep{unlockideas,novbench,llmjudge,scientificnoveltylimits,RUVERBENCH}.
As AI research systems generate ideas at scale, reliable idea reward modeling becomes crucial for assessing, selecting, and improving candidate ideas before costly research execution.

\textbf{Reward Modeling of Research Idea Generation.}
Reward models 
provide preference-based reward signals for LLM post-training~\citep{ouyang2022instructgpt,rlhflow,deepseekr1} and inference-time scaling~\citep{irvine2023rewardingchatbots,inferencescalinglaws,llmonkeys,deepseekgrm}.
Existing reward models 
primarily assess general response qualities such as helpfulness, correctness, safety, and style~\citep{rewardbench,armorm,internlm2,storyalign}.
However, research idea generation requires scientific judgment over research novelty, feasibility, and significance, which general alignment signals do not directly capture.
Recent work has begun to train research-specific evaluators through paper-review models~\citep{cycleresearcher,deepreview} or abstract preference models~\citep{scijudge}.
Yet, 
these models mainly operate at the paper level and rely on a single dominant quality signal, making them less suited to forward-looking, multifaceted idea reward modeling.
\IdeaReward instead learns from multi-signal idea-level preferences and consistently outperforms these baselines on \IdeaRewardBench.

\section{Conclusion}
This paper presents the first systematic study of reward modeling for research idea generation.
We introduce \IdeaRewardBench, a living, multi-signal benchmark for research idea evaluation with 3,056 idea preference instances, and \IdeaReward, a dedicated reward model for research idea generation.
We evaluate \IdeaReward against 23 baselines on \IdeaRewardBench and find that \IdeaReward achieves state-of-the-art performance.
These results highlight substantial room for improving idea evaluation, and encourage further research on developing stronger reward models that can reliably assess, filter, and improve the growing volume of LLM-generated research ideas.

\subsection*{AI use statement}


In this work, we used generative AI tools to support data reformatting, synthetic data generation, code editing for data processing and figure drawing, and paper polishing.
Our primary use of AI was to extract structured idea cards from openly released papers, anonymize paper-derived ideas, and generate research ideas from seed topics or partial idea fields, as described in Section~\ref{sec:2_benchmark_data_construction}.
We have not used generative AI tools to develop theoretical models or conceptual frameworks, formulate mathematical claims, propose or refine hypotheses, design or provide feedback on research  methodology or experiments, interpret results, prove mathematical claims, or assist with translation.
We have reviewed all AI-assisted outputs, including generated data, edited code, figures, and paper text, to ensure that they are accurate and consistent with our intended use.
We take responsibility for the final content of this work,
including text, claims or artifacts produced with the aid of generative AI.

\subsection*{Ethics statement}


\textbf{Intended Use.} \IdeaRewardBench and \IdeaReward are intended to support research on evaluating and improving AI-assisted research idea generation.
Through a living benchmark, a trained reward model, and comprehensive experiments, we aim to provide the academic community with diagnostic insights for developing better idea evaluator and generator models.
They are not intended to replace expert review in judging research value or making publication decisions.
Furthermore, our data and models should not be used to evaluate or improve research ideas that pose safety risks or for commercial profit.

\textbf{Potential Biases.} Since idea quality can be subjective and field-dependent, our benchmark may reflect idea preference biases present in peer-review outcomes, citation dynamics, and expert annotations. 
We therefore report results across multiple complementary idea quality signals and encourage users to treat model scores as decision-support signals rather than authoritative judgments.

\textbf{Intellectual Property.}
We construct paper-derived ideas from publicly available papers collected through OpenReview and arXiv.
Our data construction pipeline applies license screening to all paper-derived benchmark and training data.
We retain source materials licensed for redistribution and adaptation and follow the attribution requirements specified by the applicable licenses.
Appendix~\ref{sec:appendix_data_licensing} describes the detailed license screening procedure.


\subsection*{Reproducibility statement}


We make our best efforts to ensure that results in this work are reproducible. 
The construction of \IdeaRewardBench is specified in Section~\ref{sec:2_benchmark_data_construction},
where the data construction, training objective, and training hyperparameters for \IdeaReward are described in
Sections~\ref{sec:3_training_data_construction} and~\ref{sec:3_training_ideareward}.
More details on the idea extraction and anonymization pipeline, \IdeaReward training data construction, data licensing and attribution, evaluation protocol, baseline adaptation procedures, and LLM prompts are provided in Appendices~\ref{sec:appendix_llm_idea_extraction}, \ref{sec:appendix_training_data_construction}, \ref{sec:appendix_data_licensing}, \ref{sec:appendix_evaluation_details}, \ref{sec:appendix_baseline_adaptations}, and \ref{sec:appendix_prompts}. 
Upon publication, we will release \IdeaRewardBench evaluation data, \IdeaReward training data, and \IdeaReward model checkpoints, along with code for automated idea extraction and anonymization, reward-model training, and benchmark evaluation.


\bibliography{iclr2027_conference}
\bibliographystyle{iclr2027_conference}

\newpage
\appendix
\section{Correlation between idea quality and paper quality}
\label{sec:appendix_preliminary_experiment}
For the review-grounded and impact-grounded splits, \IdeaRewardBench uses paper-level outcomes as proxies for the extracted research ideas.
We conduct a pilot validation on 200 instances sampled from the benchmark construction pool, comprising 50 review-grounded pairs, 50 review-grounded quadruples, 50 impact-grounded pairs, and 50 impact-grounded quadruples.
Two human annotators independently rank the anonymized idea cards and resolve disagreements through discussion.
We then compare the final human idea-level preferences with the gold paper-level preferences derived from conference outcomes or citation impact.

\begin{table}[h]
\centering
\small
\begin{tabular}{lcc}
\toprule
Split & $N$ & Idea--paper agreement \\
\midrule
Review-grounded pairs & 50 & 88.0\% \\
Review-grounded quadruples & 50 & 84.0\% Top-1 / 90.3\% PRA \\
Impact-grounded pairs & 50 & 84.0\% \\
Impact-grounded quadruples & 50 & 82.0\% Top-1 / 92.0\% PRA \\
\midrule
Overall & 200 & 84.5\% top-choice agreement \\
\bottomrule
\end{tabular}
\caption{Agreement between paper-level preferences and blind human preferences over anonymized idea cards.}
\label{tab:paper_idea_quality_correlation}
\end{table}

Before disagreement resolution, the two annotators achieve 80.5\% overall top-choice agreement, with Cohen's $\kappa=0.710$.
These results support paper-level outcomes as informative and scalable, though imperfect, proxies for comparative idea quality.

\section{Example Data of \IdeaRewardBench}

In this section, we show example instances of \IdeaRewardBench in the original JSON structure.

\begin{tcblisting}{
  title={Review-grounded quadruple example},
  colback=gray!2,
  colframe=gray!55!black,
  fonttitle=\bfseries\footnotesize,
  boxrule=0.6pt,
  arc=1.2pt,
  breakable,
  listing only,
  listing options={basicstyle=\ttfamily\tiny, breaklines=true, columns=fullflexible}
}
{
  "split": "review-grounded-extracted-quadruple",
  "metadata": {
    "venue": "iclr2026",
    "primary_area": "generative models",
    "search_cutoff_date": "2025-09-04"
  },
  "candidates": [
    {
      "final_decision": "strong_accept",
      "idea_card": {
        "title": "Aligning Diffusion Language Model Training with Non-Uniform Inference Planners",
        "research_question": "How should the training of denoisers in diffusion language models be adapted when inference inevitably proceeds under a non-uniform planner, given that standard training assumes uniformly random denoising paths?",
        "motivations": "Diffusion language models enable flexible, parallel generation by iteratively denoising masked sequences, but their standard training objective assumes positions are unmasked uniformly at random. In practice, inference uses planners (e.g., greedy decoding, confidence-based sampling) that select which tokens to reveal next based on confidence, substantially improving sample quality across text, code, and protein domains. However, this creates an irrevocable mismatch: the forward masking process used during training corrupts positions uniformly, while the reverse process at inference follows a planner-driven, non-uniform path. It can be proven that under an imperfect denoiser, the standard evidence lower bound does not accurately bound the log-likelihood under greedy ancestral sampling, meaning the training objective no longer reflects the quality of planner-generated samples. This mismatch matters because the denoiser is effectively trained on a harder, uniform-unmasking problem rather than the planned generation paths it will actually encounter, degrading sample quality.",
        "methods": "The proposed study would derive a planner-aware evidence lower bound that incorporates planner-based reverse dynamics directly into the training objective. The framework would model diffusion language model sampling as a discrete-time Markov chain and introduce a planner function that selects which masked coordinate to denoise next, decomposing each reverse step into: (1) the denoiser producing candidate predictions for all positions, and (2) the planner sampling an index to unmask. The derived bound would consist of two terms: a planner-weighted cross-entropy (log-probability of correct tokens weighted by planner selection probability) and a correction term measuring the KL divergence between an ideal planner with ground-truth access and the effective planner relying on denoiser predictions. For the uniform planner, the correction term would vanish, recovering the standard evidence lower bound. The practical algorithm would use a soft greedy planner that relaxes the deterministic argmax into a softmax over denoiser confidence with a temperature parameter. Gradients would be detached through the planner to remove the expensive correction term, leaving only the weighted cross-entropy. To stabilize training, the method would interpolate between the vanilla uniform loss and the planner-weighted loss, replacing uniform weights with planner-adjusted weights controlled by a planner weighting strength parameter. The masking scheme would reuse the vanilla uniform random masking (no path simulation needed), making the approach a simple modification to the standard masked diffusion loss with no additional computational overhead.",
        "experiment_plan": "The evaluation would assess the proposed method across three domains: protein sequence generation, unconditional text generation, and code generation (completion and infilling). For protein generation, a baseline and the proposed variant would be trained under identical configurations with confidence-based sampling at inference; generated sequences would be folded with ESMFold and evaluated on pLDDT, pTM, pAE, foldability, entropy, and diversity, compared against existing protein generation baselines. For text generation, models would be trained on OpenWebText and evaluated with MAUVE, generative perplexity, and entropy across varying sampling steps, compared against autoregressive and diffusion baselines. For code generation, models initialized from Qwen2.5-Coder would be trained on a code dataset and evaluated on HumanEval, HumanEval+, MBPP, MBPP+, and code infilling benchmarks with pass@k and exact match, compared against autoregressive and diffusion baselines. Ablations would study the softmax temperature and path learning weight. Additional analyses would include training convergence curves, robustness to sampling steps and temperature variation (from scratch and fine-tuning), and comparison of training stability between the pure proposed loss and the interpolated loss."
      }
    },
    {
      "final_decision": "standard_accept",
      "idea_card": {
        "title": "Exploiting Directional Curvature of Log-Probabilities for Denoising-Free Memorization Detection and Mitigation in Diffusion Models",
        "research_question": "Can the directional variation (anisotropy) of the log-probability in the low-noise regime of diffusion models be leveraged to overcome the limitations of norm-based memorization detection metrics, and can a combined isotropic-anisotropic metric enable more effective and efficient denoising-free memorization detection and mitigation?",
        "motivations": "Diffusion models are vulnerable to memorization, where they reproduce exact or near-replica training images, raising data privacy and copyright concerns. Existing norm-based memorization detection methods approximate the overall curvature of the log-probability, which is effective only under the assumption of isotropic log-probability distributions that hold at high or medium noise levels. However, in the low-noise anisotropic regime, curvature varies by direction, meaning a high norm in one direction can be compensated by a low norm in another, causing norm-based metrics to fail and potentially leading to false negatives. This limitation is particularly problematic in scenarios like image-level memorization detection, where generated images reside far from the high-noise isotropic regime and directly on the data manifold. Thus, a detection method that accounts for directional curvature is needed to reliably characterize memorization across noise levels.",
        "methods": "The proposed study would develop a denoising-free memorization detection metric that combines isotropic and anisotropic signals of the log-probability. The core insight is that in the anisotropic low-noise regime, memorization manifests as strong angular alignment between the guidance vector (the conditioning term, i.e., the difference between conditional and unconditional scores) and the unconditional score estimate. The metric would be formulated as a weighted sum of two components: (a) the norm of the guidance vector in the isotropic high-noise regime (capturing overall curvature), and (b) the cosine similarity between the guidance vector and the unconditional score estimate in the anisotropic low-noise regime (capturing directional alignment). The metric would be computed directly on pure noise inputs using only two forward passes of the model: one conditional and one unconditional pass at a high-noise timestep (t approximately T) for the isotropic component, and one at a low-noise timestep (t approximately 0) for the anisotropic component, eliminating costly denoising steps. For memorization mitigation, the metric would be integrated into a prompt augmentation scheme that adapts memorized prompts during inference to steer generation away from memorized samples.",
        "experiment_plan": "The evaluation would cover two tasks: memorization detection and inference-time mitigation. Detection experiments would be conducted on Stable Diffusion v1.4 and v2.0 using 500 and 219 memorized prompts respectively, alongside 500 non-memorized prompts from Lexica, GPT-4, COCO, and Tuxemon. Metrics would include AUC, TPR@1%FPR, and computation time for 10 prompts, evaluated for n=1 and n=4 generations. Baselines would include denoising-free methods by Ren et al. (2024), Wen et al. (2024), and Jeon et al. (2025). Mitigation experiments would use a memorization benchmark suite, evaluating SSCD Similarity Score, CLIP Score, and Aesthetic Score across five hyperparameter configurations, compared against no mitigation and prior mitigation methods. Ablation studies would compare alternative cosine similarity formulations, assess the individual contributions of the isotropic norm and anisotropic alignment components, and analyze normalization and timestep choices. Additional generalization experiments would be performed on Realistic Vision."
      }
    },
    {
      "final_decision": "standard_reject",
      "idea_card": {
        "title": "Training-Free, Robust Watermarking Across Diverse Visual Synthesis Paradigms",
        "research_question": "How can a general-purpose, reliable, and robust watermarking method be designed for diverse vision generative paradigms (diffusion, autoregressive, and hybrid) without requiring model-specific fine-tuning or sacrificing perceptual quality?",
        "motivations": "Designing a general-purpose watermarking method for vision generative models is challenging due to the diversity of modern generative paradigms, each with distinct architectures, training objectives, and inference procedures. Existing approaches suffer from three main limitations: (a) paradigm-specific techniques that cannot transfer across frameworks, (b) paradigm-agnostic post-hoc strategies that are vulnerable to image transformations, and (c) fine-tuning-based methods whose effectiveness depends heavily on generalization ability. Given that generative paradigms have not yet converged and new frameworks continue to emerge, there is a critical need for a general watermarking algorithm that enables plug-and-play injection, supports reliable detection, remains robust against image transformations, and preserves perceptual quality.",
        "methods": "The proposed approach would introduce a watermark definition based on patch-level luminance statistics and a training-free injection algorithm using a guidance mechanism. For the watermark definition, the service provider would predefine a binary pattern and corresponding patch-level thresholds. To detect a watermark, the image would be partitioned into patches, the luminance of each patch computed, and detection would check whether the sign of (luminance minus threshold) matches the predefined binary pattern. A statistical guarantee would show that for unwatermarked images, the false positive rate decreases exponentially with the number of patches. For injection, the method would leverage the widely adopted guidance technique as a plug-and-play mechanism across diverse generative paradigms. The guidance signal would direct each generation step toward alignment with the predefined pattern by softly injecting the watermark constraint into the generation process, adjusting content smoothly to preserve image quality. If the initial generation does not exceed the match rate threshold, images would be repeatedly generated until the constraint is satisfied.",
        "experiment_plan": "The evaluation would be conducted on nine ImageNet-pretrained generative models spanning diffusion (EDM2-XS, EDM2-L, EDM2-XXL), autoregressive (VAR-d16, VAR-d30, VAR-d36), and hybrid (MAR-Base, MAR-Large, MAR-Huge) frameworks, covering different resolutions and parameter scales. Robustness would be evaluated by sampling 1,000 unwatermarked and 1,000 watermarked images per method, applying nine image transformations (Scaling, Cropping, JPEG Compression, Filtering, Smoothing, Color Jitter, Color Quantization, Gaussian Noise, Sharpening), and measuring detection accuracy. Baselines would include post-hoc methods (DwtDct, DwtDctSvd, RivaGAN) and diffusion-specific methods (Gaussian-Shading, PRC-W). Fidelity would be measured using Frechet Inception Distance (FID) on 50,000 watermarked images against the clean dataset, comparing against Gaussian-Shading and PRC-W. Ablation studies would compare the proposed guidance against post-hoc projection and hard step-wise projection on EDM2-L, VAR-d30, and MAR-Large, and compare the injection with alternative watermark patterns (random, RGB average). All experiments would be repeated 20 times with different random seeds."
      }
    },
    {
      "final_decision": "strong_reject",
      "idea_card": {
        "title": "Embedding Dynamic Causal Knowledge into Diffusion Models for Cross-Modal Physiological Waveform Generation",
        "research_question": "How can dynamic, sample-specific causal knowledge governing cardiovascular dynamics be explicitly embedded into the reverse diffusion process of a DDPM to ensure that synthesized physiological signals remain physiologically plausible, particularly under out-of-distribution conditions such as arrhythmias?",
        "motivations": "While Denoising Diffusion Probabilistic Models (DDPMs) excel at generating statistically plausible signals, their purely data-driven paradigm often lacks consistency with known physiological processes, producing visually convincing but physiologically implausible artifacts (e.g., a dicrotic notch preceding the systolic peak). Existing methods for incorporating prior knowledge, such as Physics-Informed Neural Networks (PINNs), typically rely on global constraints or simplified conditioning schemes that lack the granularity to explicitly model the multi-dimensional, dynamic causal relationships governing complex physiological systems. This limitation is particularly critical in physiological signal synthesis, where the interplay between different modalities (e.g., ECG, PPG, and ABP) is governed by highly dynamic and nonlinear causal mechanisms. A capability gap exists for generative models that can move beyond statistical fidelity to incorporate domain-specific causal knowledge as structured regularizers, ensuring physiological plausibility and robustness when statistical patterns are disrupted.",
        "methods": "The proposed approach would augment a standard conditional DDPM by conditioning the U-Net denoiser not only on the diffusion timestep and multi-modal input signals but also on a learned causal graph. The core would be a causal structure builder operating in two stages. First, an offline knowledge fusion framework would construct a static base causal graph by combining a prior graph derived from established physiological domain knowledge with a data-driven graph discovered by the Fast Causal Inference (FCI) algorithm; the fusion would be performed via an element-wise logical OR of their adjacency matrices. Second, an online dynamic graph adapter would transform this static skeleton into a sample-specific guide: a multilayer perceptron (MLP) would project pooled input node features into a dynamic modulation matrix, which would be multiplied (Hadamard product) with the static base adjacency matrix to produce a dynamic, sample-specific weighted adjacency matrix. A Graph Convolutional Network (GCN) would then process this dynamic graph to generate node embeddings, which would be aggregated into a causal vector that modulates the U-Net denoiser at multiple feature scales. The model would be trained end-to-end with a multi-objective loss combining a hybrid L1/L2 diffusion loss for noise prediction and an auxiliary MSE loss that supervises a physiological feature estimator to estimate physiological components from the input signals, grounding the causal graph embedding in accurate physiological estimations.",
        "experiment_plan": "The framework would be validated on two cross-modal tasks: continuous ABP waveform generation from ECG and PPG inputs, and PPG-to-ECG synthesis. Datasets would include the MIMIC-III public dataset (84,110 segments of synchronized PPG, ECG, and ABP at 125 Hz) and a proprietary arrhythmia dataset (36,537 cardiac cycles from 48 patients at 250 Hz). Data would be split 8:1:1 by patient to prevent leakage. Baselines would span multiple architectural paradigms including BiLSTM, CNN, WaveNet, UNet, GSW-UNet, Transformer, Mamba, RDAE, KD-Informer, and a standard conditional DDPM. Evaluation would use waveform fidelity metrics (RMSE, MAE, Pearson Correlation) and clinical parameter accuracy (ME and SDE for SBP/DBP against the AAMI standard for ABP; HR, RMSSD MAE, QRS Sensitivity, and PPV for ECG). An ablation study would isolate the contribution of the causal guidance module by comparing the proposed model against a standard conditional DDPM. Robustness would be specifically assessed under pathological conditions using the arrhythmia dataset, and qualitative analysis would examine generated waveforms for physiological plausibility (e.g., correct temporal ordering of systolic peak and dicrotic notch)."
      }
    }
  ]
}
\end{tcblisting}

\begin{tcblisting}{
  title={Impact-grounded quadruple example},
  colback=gray!2,
  colframe=gray!55!black,
  fonttitle=\bfseries\footnotesize,
  boxrule=0.6pt,
  arc=1.2pt,
  breakable,
  listing only,
  listing options={basicstyle=\ttfamily\tiny, breaklines=true, columns=fullflexible}
}
{
  "split": "impact-grounded-extracted-quadruple",
  "metadata": {
    "venue": "arxiv",
    "search_cutoff_date": "2025-03-30",
    "license_audit": "all source papers CC BY 4.0"
  },
  "candidates": [
    {
      "impact_rank": 1,
      "citation_count": 503,
      "idea_card": {
        "title": "Scaling Reinforcement Learning on Base Language Models via Minimalist Policy Optimization with Learned Critics for Reasoning",
        "research_question": "Can a minimalist reinforcement learning approach using clipped policy optimization with generalized advantage estimation (lambda=1, gamma=1), rule-based rewards, and no KL regularization achieve stable and scalable reasoning-oriented training directly on a base language model, and how does a learned critic improve advantage estimation over group-relative policy gradient methods for this setting?",
        "motivations": "Large-scale reinforcement learning training of language models on reasoning tasks has shown promising scaling phenomena, as benchmark performance and response length steadily increase with more training computation. However, prior open implementations of this training paradigm have been limited, and group-relative policy gradient methods that lack a dedicated value network struggle to distinguish genuinely correct responses from those occurring within negative patterns such as repetitive loops, which can misdirect reinforcement and lead to training instability or collapse. There is a need for an accessible, reproducible, and robust training recipe that overcomes value estimation instability and enables steady scalability across model sizes.",
        "methods": "The approach would apply large-scale reinforcement learning training directly to a base language model (e.g., Qwen2.5-32B base). The core algorithm would be clipped policy optimization with generalized advantage estimation using lambda=1 and gamma=1, which simplifies the advantage computation to the terminal reward minus the learned value function, where the reward is a single terminal rule-based signal (0 or 1) and the value function is learned during training. The policy would be updated via the standard clipped surrogate objective with epsilon=0.2, and the value network would be trained to minimize squared error against the terminal reward. No KL regularization (neither KL loss nor KL penalty) would be used, allowing unconstrained exploration away from the base model distribution. A rule-based reward function would assign a single terminal reward at the end of each generated trajectory based on answer correctness. The learned critic would provide token-level value estimation that identifies and devalues repetitive or degenerate patterns, yielding more robust advantage estimates compared to group-relative policy gradient methods. Training would be conducted on a carefully curated 57k dataset spanning arithmetic, logic, coding, and common-sense reasoning, with data preprocessing to handle diverse answer formats.",
        "experiment_plan": "The evaluation would train models at multiple scales (0.5B, 1.5B, 7B, 32B) using the Qwen2.5 base models and assess performance on AIME 2024, AIME 2025, MATH500, and GPQA Diamond benchmarks, comparing against existing open reinforcement-learning-trained baselines. Ablation studies would investigate: (1) generalized advantage estimation lambda values (1.0 vs. 0.95) for training stability and response length dynamics; (2) KL regularization (no KL vs. KL loss vs. KL reward shaping) for performance and computational efficiency; (3) data scale (the curated 57k dataset vs. MATH train 7.5k) for sustained improvement in reward and response length. The study would analyze the learned critic and advantage estimation by comparing advantage assignments from the proposed method to group-relative policy gradient methods on tokens with repetitive patterns, visualizing value approximations for coherent vs. repetitive text. Training dynamics including correct format ratio and reflection pattern lengths would be analyzed. The pipeline would be extended to smaller models (0.5B, 1.5B) and to distilled models to test generalization."
      }
    },
    {
      "impact_rank": 2,
      "citation_count": 97,
      "idea_card": {
        "title": "Probabilistic Curriculum Scheduling for Reinforcement Learning Post-Training of LLM Reasoning",
        "research_question": "Can scheduling reinforcement learning post-training tasks from easy to hard, using a probabilistic scheduler that gradually fades out easier tasks, enable LLMs to learn reasoning skills on inherently difficult tasks where direct RL fails due to sparse rewards and distribution gaps?",
        "motivations": "Prior RL post-trained models have shown reasoning abilities, but using RL alone to improve reasoning on inherently difficult tasks is less effective because rewards are sparse and the base model has low zero-shot performance. A non-trivial distribution gap exists between the model's pre-training data distribution and the target hard task distribution, leading to low accuracy and sparse reward signals. Additionally, fitting the model to a single target distribution can cause overfitting and memorization, undermining generalization. Prior curriculum learning efforts for LLM post-training rely on simplified strategies such as fixed-iteration switching, which can cause task forgetting and overfitting to trivial tasks. There is a need for a method that builds reasoning skills gradually while preventing reward hacking and forgetting.",
        "methods": "The proposed approach formulates LLM reasoning as a discounted MDP where states are token prefixes, actions are vocabulary tokens, and the policy is the pre-trained LLM. It uses sparse rewards: zero for intermediate states and a correctness-based reward only when the final answer wrapped in designated answer tags is completed. The method first decomposes the training dataset into four difficulty levels (trivial, easy, medium, hard) using either human-aligned signals (plan length, labeled difficulty, operand count) or automatically estimated difficulty based on model error rates under 1-shot chain-of-thought prompting with 20 responses per question, grouped by quartiles. A probabilistic scheduler then gradually shifts sampling focus from easy to hard tasks. Two schedulers would be developed: (1) a cosine scheduling variant, a non-parametric strategy where sampling probability for each task is set by ordinal number and interpolated via a cosine function, and (2) a Gaussian scheduling variant, inspired by Gaussian mixture models, where sampling probability follows Gaussian distributions with means at task indices and shared variance, with a position parameter that determines how quickly the focus moves to harder tasks. The RL algorithm used would be GRPO. Theoretical analysis under an approximate policy iteration framework would establish convergence guarantees and finite-sample complexity bounds, examining whether curriculum stages require fewer total samples than direct learning when tasks are appropriately decomposed.",
        "experiment_plan": "The evaluation would be conducted on Blocksworld, Countdown, MATH, GSM8K, and AQuA using Qwen 2.5 1.5B Instruct and LLaMA 3.2 3B Instruct. Datasets with human-annotated difficulty (Blocksworld by plan length, Countdown by number of operations, MATH by problem levels) would use provided labels, while GSM8K and AQuA would use zero-shot error rate quartiles. An out-of-distribution split would be defined for datasets with difficulty labels. Baselines would include chain-of-thought prompting, GRPO with balanced scheduling, GRPO trained only on hard or OOD subsets, traditional curriculum learning, self-evolution methods, and SFT. Metrics would include per-difficulty-level accuracy and pass@k evaluation. Ablations would examine the role of task decomposition, the impact of different schedulers (cosine vs. Gaussian with varying hyperparameters), and compatibility with DAPO as an alternative RL algorithm. Additional analysis would include pass@k on AIME24 and OlympiadBench for models trained on MATH, robustness to the number of difficulty splits, and comparison of Gaussian scheduling parameter settings."
      }
    },
    {
      "impact_rank": 3,
      "citation_count": 24,
      "idea_card": {
        "title": "Dynamic Sample Budget Allocation for Chain-of-Thought Reasoner Training via Gradient Variance Minimization",
        "research_question": "How can the stochastic gradient variance in chain-of-thought training be minimized under a fixed computational budget, given that uniform best-of-n sampling in iterative reward-ranked fine-tuning methods fails to account for prompt-specific difficulty and convergence behavior?",
        "motivations": "Chain-of-thought reasoning in LLMs can be formalized as a latent variable problem solved via EM-style iterative reward-ranked fine-tuning, where the E-step approximates the posterior over rationales through rejection (best-of-n) sampling. Prior approaches apply uniform inference budgets across all prompts, which ignores variability in prompt difficulty and convergence behavior. This static allocation biases accepted samples toward easier prompts with higher acceptance rates, while difficult prompts with low acceptance rates introduce high gradient variance and receive insufficient useful training signal. The key bottleneck is inefficient Monte Carlo estimation of the ELBO gradient during the E-step, motivating a principled dynamic sample allocation strategy that minimizes stochastic gradient variance under a computational budget constraint.",
        "methods": "The proposed method would refine iterative reward-ranked fine-tuning with dynamic inference budget scheduling via gradient variance minimization. The approach would formalize CoT as prompt->rationale->answer and derive an unbiased gradient estimator for the EM objective, where the variance of the estimator depends on per-prompt acceptance rate and gradient norm. The algorithm would operate in two stages per iteration: (1) Estimation stage--for each prompt, sample a small number of times to estimate acceptance rate and a Lipschitz coefficient computed as the norm of gradients of log-likelihood for accepted responses; (2) Budget Assignment--solve a regularized optimization problem minimizing the sum of squared gradient norms divided by per-prompt sample count and acceptance rate, weighted by a penalty term that prevents excessive sampling on uninformative prompts with near-zero acceptance rates, subject to a total budget constraint. A closed-form solution would allocate samples proportional to gradient norm divided by the square root of a regularized acceptance rate term. The M-step would then perform gradient descent using the allocated samples. The dynamic allocation strategy could be disentangled from the EM pipeline and applied to reinforcement learning algorithms by rebalancing rollout budgets across prompts. Hyperparameters controlling the penalty and regularization would be tuned; the posterior and budget would be updated periodically across M-steps.",
        "experiment_plan": "The evaluation would use Qwen2.5-Math-1.5B and Qwen2.5-Math-7B on mathematical reasoning with Math-Verify as the symbolic verifier and Numina-Math as the training dataset. The method would be integrated into both iterative reward-ranked fine-tuning and reinforcement learning fine-tuning pipelines using the verl training framework. Evaluation would be on five benchmarks: Math500, Minerva Math, Olympiad Bench, AIME24, and AMC23, measured by Average@8 accuracy. Baselines would include vanilla iterative reward-ranked fine-tuning and reinforcement learning fine-tuning with uniform per-prompt rollout of 8. Ablations would vary the estimation sample size (8, 16, 32) and total budget (4n, 8n, 16n, 32n), and analyze convergence rate and final accuracy. Additional ablations would examine the choice of gradient-calculation layers (lm_head vs. embed_tokens), gradient reduction methods (summation vs. average by sequence length), and the effects of penalty and regularization hyperparameters on sample weight allocation. Step-wise convergence curves would compare the dynamic allocation variant against uniform allocation baselines."
      }
    },
    {
      "impact_rank": 4,
      "citation_count": 7,
      "idea_card": {
        "title": "Episodic-Memory-Augmented Reinforcement Learning for Chain-of-Thought Reasoning in Sub-Billion-Parameter Language Models",
        "research_question": "How can reinforcement learning fine-tuning be made effective for tiny LLMs (<=1B parameters) for chain-of-thought reasoning, given that these models suffer from extremely sparse outcome rewards, lack effective exploration mechanisms, and are prone to training collapse when relying on format-based heuristic rewards?",
        "motivations": "RL fine-tuning has improved reasoning in large LLMs (8B-670B), but tiny LLMs (<=1B) lack the pretraining strength to explore effectively, frequently producing incorrect or improperly formatted outputs that yield zero outcome reward, making reward signals extremely sparse. A common mitigation--format-based heuristic rewards--can cause training collapse in tiny LLMs, as they overfit to simple format patterns while neglecting the main task. Additionally, LLMs passively sample from learned distributions without an explicit exploration or exploitation mechanism, which is especially problematic for small models that choose poor actions. There is also a lack of quality data in downstream tasks for training tiny LLMs with RL. These limitations motivate a method that provides dense, performance-driven learning signals and an explicit exploration/exploitation mechanism tailored to tiny LLMs.",
        "methods": "The proposed method would augment RL fine-tuning (specifically group-relative policy optimization) with a kNN-based episodic memory that stores past query-response pairs and computes intrinsic rewards balancing exploration and exploitation. Two separate episodic memory modules would be maintained: a success memory and a failure memory. Queries and responses would be encoded into a shared vector space using a pre-trained Sentence Transformer. During training, G responses would be sampled per query; responses would be classified by outcome reward thresholds and written into the success or failure memory, with FIFO eviction at capacity N (queries) and L (responses per query). For a new query and generated response, the top-k nearest queries would be retrieved from each memory via cosine similarity. The exploitation reward would be the negative Euclidean distance between the response embedding and the centroid of retrieved success response embeddings, encouraging alignment with generalizable successful patterns. The exploration reward would be 1 minus the maximum cosine similarity between the response embedding and retrieved failure response embeddings, penalizing similarity to past mistakes. Both intrinsic rewards would be normalized via running min-max scaling over a sliding window of size w. The final intrinsic reward would be a weighted sum of the normalized exploitation and exploration components, added to the outcome reward from an answer verifier, and the policy would be optimized via group-relative policy optimization. A warm-up period (e.g., 50 steps) would collect initial data before applying the exploration reward.",
        "experiment_plan": "The evaluation would fine-tune three tiny LLMs--Qwen2.5-0.5B-Instruct, Falcon3-1B-Instruct, and Llama3.2-1B-Instruct--on GSM8K (7,473 training samples) and a 2,000-sample subset of the AI-MO dataset, using a single NVIDIA H100 and three random seeds. Evaluation would be conducted on GSM8K, MATH-500, and AIME24 under zero-shot settings using Lighteval's extractive match metric. Baselines would include a standard group-relative-policy-optimization baseline with correctness and format-based rewards, a response-length-based reward baseline, an exploitation-only memory variant, and the full method with both exploitation and exploration rewards. Key hyperparameters: G=16 generations per step, max response length 200, memory capacity N=|D|, L=100 responses per query, normalization window w=100, K=1 for the main GSM8K experiments. Ablations would study the effect of K (1, 10, 20, 30) on AI-MO fine-tuning of Qwen2.5-0.5B-Instruct. Analysis experiments would investigate training collapse phenomena (reward mode collapse, response length collapse), response diversity using TF-IDF and Sentence Transformer embeddings on 100 geometry questions, and emergence of self-verification behaviors in generated outputs."
      }
    }
  ]
}
\end{tcblisting}

\begin{tcblisting}{
  title={Annotation-grounded human-vs-LLM pair example},
  colback=gray!2,
  colframe=gray!55!black,
  fonttitle=\bfseries\footnotesize,
  boxrule=0.6pt,
  arc=1.2pt,
  breakable,
  listing only,
  listing options={basicstyle=\ttfamily\tiny, breaklines=true, columns=fullflexible}
}
{
  "split": "annotation-grounded-human-vs-llm-rq-pair",
  "metadata": {
    "venue": "iclr2026",
    "primary_area": "foundation or frontier models, including LLMs",
    "search_cutoff_date": "2025-09-19",
    "research_question": "Can we cast reward modeling as a reasoning task?",
    "motivations": "Existing scalar-based reward models are opaque, offering no intermediate reasoning steps to justify their decisions, which limits their capacity to handle reasoning-intensive preference tasks. While generative reward models provide greater transparency, their reasoning is often superficial and unhelpful for reliable judgment, leading to suboptimal performance. Accurate and grounded reward modeling in real-world scenarios requires jointly conducting reasoning and reward assignment, as preference judgments inherently involve multifaceted cognitive considerations such as inferring latent evaluation criteria and navigating trade-offs among multiple criteria. Prior critique-based generative methods have relied heavily on unstructured, self-generated chain-of-thought reasoning from instruct models, limiting their reasoning capabilities and resulting in inferior performance compared to scalar counterparts."
  },
  "candidates": [
    {
      "final_decision": "standard_accept",
      "source": "human",
      "idea_card": {
        "title": "Reasoning-Driven Reward Modeling via Distillation and Reinforcement Learning",
        "research_question": "Can reward modeling be effectively cast as a reasoning task, where the model generates structured evaluation criteria and step-by-step justifications before assigning preference judgments?",
        "motivations": "Existing scalar-based reward models are opaque, offering no intermediate reasoning steps to justify their decisions, which limits their capacity to handle reasoning-intensive preference tasks. While generative reward models provide greater transparency, their reasoning is often superficial and unhelpful for reliable judgment, leading to suboptimal performance. Accurate and grounded reward modeling in real-world scenarios requires jointly conducting reasoning and reward assignment, as preference judgments inherently involve multifaceted cognitive considerations such as inferring latent evaluation criteria and navigating trade-offs among multiple criteria. Prior critique-based generative methods have relied heavily on unstructured, self-generated chain-of-thought reasoning from instruct models, limiting their reasoning capabilities and resulting in inferior performance compared to scalar counterparts.",
        "methods": "The proposed approach would formulate reward modeling as a reasoning task using a two-stage training pipeline: reasoning distillation and reinforcement learning (RL). In the distillation stage, starting from an off-the-shelf instruction-tuned model (e.g., Qwen-2.5-Instruct), the model would be trained using synthesized high-quality reasoning traces to bootstrap reasoning ability. In the RL stage, the model would be further optimized using RL with verifiable rewards to improve generalization beyond distilled patterns. The approach would feature a structured rubric generation mechanism with query type classification: the model would categorize the input sample as either chat or reasoning. For chat tasks, the model would generate sample-specific evaluation rubrics, justifications for the rubrics, and evaluations tailored to the question. For reasoning tasks, the model would first solve the problem itself before evaluating and picking the preferred response. For existing reasoning models that have already undergone reasoning-focused distillation (e.g., DeepSeek-R1-Distill-Qwen), the model would be fine-tuned using RL with verifiable rewards without the additional distillation stage. The RL objective would use GRPO with verifiable rewards based on whether the model's final judgment matches the ground-truth preference label.",
        "experiment_plan": "The proposed approach would be evaluated on three primary benchmarks: RewardBench, RM-Bench, and RMB, covering categories such as chat, chat-hard, reasoning, safety, math, and code. Training would utilize a cleaned subset of Skywork Reward Preference 80K (excluding magpie_ultra data due to spurious correlations), 8K examples from Code-Preference-Pairs, and the full Math-DPO-10K dataset. Baselines would include scalar reward models (e.g., INF-ORM-Llama3.1-70B, Nemotron-4-340B-Reward), generative reward models (e.g., GPT-4o, Claude-3-5-sonnet, Skywork-Critic-Llama-3.1-70B), and other reasoning-based reward models (e.g., JudgeLRM, DeepSeek-GRM). Ablation studies would compare settings including cold-start RL, cold-start RL with structured rubric generation, cold-start RL with structured rubric generation and query type classification, and the full distillation plus RL plus rubric generation plus query classification pipeline. Additional analyses would examine model size scaling (7B, 14B, 32B based on Qwen-2.5-Instruct), inference-time compute scaling (512 to 8192 tokens), reasoning training effectiveness versus SFT-only approaches, training dynamics comparing cold-start versus warm-start RL, and qualitative case studies of rubric generation and evaluation behavior."
      }
    },
    {
      "final_decision": "standard_reject",
      "source": "llm",
      "llm_name": "Qwen3.5-35B-A3B",
      "idea_card": {
        "title": "Structured Reasoning Reward Models: Explicit Criteria Discovery for Transparent Preference Learning",
        "research_question": "Can reward modeling be effectively cast as a reasoning task, where the model generates structured evaluation criteria and step-by-step justifications before assigning preference judgments?",
        "motivations": "Existing scalar-based reward models are opaque, offering no intermediate reasoning steps to justify their decisions, which limits their capacity to handle reasoning-intensive preference tasks. While generative reward models provide greater transparency, their reasoning is often superficial and unhelpful for reliable judgment, leading to suboptimal performance. Accurate and grounded reward modeling in real-world scenarios requires jointly conducting reasoning and reward assignment, as preference judgments inherently involve multifaceted cognitive considerations such as inferring latent evaluation criteria and navigating trade-offs among multiple criteria. Prior critique-based generative methods have relied heavily on unstructured, self-generated chain-of-thought reasoning from instruct models, limiting their reasoning capabilities and resulting in inferior performance compared to scalar counterparts.",
        "methods": "The proposed approach would develop a Structured Reasoning Reward Model (SRRM) that decomposes preference evaluation into three explicit reasoning stages.\nFirst, a Criteria Discovery Head would generate a fixed-length latent vector representing relevant evaluation dimensions specific to the prompt context.\nSecond, a Justification Generator would produce step-by-step rationales linking input tokens to the discovered criteria, constrained by a logical consistency loss.\nThird, an Aggregation Module would compute a scalar reward from the reasoning traces rather than direct token probability, ensuring the score is grounded in the preceding logic.\nTraining would utilize a multi-task objective combining pairwise ranking loss with a reasoning regularization term that penalizes logical contradictions between justification tokens and final scores.\nThe architecture would employ a shared Transformer encoder for input processing, followed by task-specific decoders for each reasoning stage, sharing parameter space to maintain efficiency.",
        "experiment_plan": "Evaluation would be conducted on HH-RLHF and UltraFeedback datasets using standard pairwise preference prediction tasks.\nPrimary metrics would include pairwise accuracy against human annotations, Kendall Tau correlation, and reasoning fidelity measured via LLM-as-a-judge scoring on the generated rationales.\nBaselines would include standard scalar Reward Models, unstructured Chain-of-Thought reward models, and direct preference optimization policy gradients.\nAdditional experiments would test robustness against adversarial prompts designed to manipulate standard reward models without affecting reasoning chains.\nAblation studies would isolate the impact of the Criteria Discovery Head and the Logical Consistency Loss by comparing performance against versions without structured reasoning outputs.\nFurthermore, the proposed model would be integrated into a standard RLHF pipeline to evaluate downstream alignment quality on open-ended generation benchmarks such as MT-Bench, comparing fluency and helpfulness scores against models trained with opaque reward signals."
      }
    }
  ]
}
\end{tcblisting}

\begin{tcblisting}{
  title={Annotation-grounded LLM-vs-LLM pair example},
  colback=gray!2,
  colframe=gray!55!black,
  fonttitle=\bfseries\footnotesize,
  boxrule=0.6pt,
  arc=1.2pt,
  breakable,
  listing only,
  listing options={basicstyle=\ttfamily\tiny, breaklines=true, columns=fullflexible}
}
{
  "split": "annotation-grounded-llm-vs-llm-topic-pair",
  "metadata": {
    "search_cutoff_date": "2026-09-13",
    "topic": "LLM Creative Generation: Methods for evaluating or improving LLM performance on creative generation tasks."
  },
  "candidates": [
    {
      "final_decision": "standard_accept",
      "source": "llm",
      "llm_name": "GLM-5.2",
      "idea_card": {
        "title": "StoryPlan: Goal-Conditioned MCTS for Long-Horizon Narrative Coherence and Twist Generation",
        "research_question": "Can a Monte Carlo Tree Search algorithm guided by a narrative surprise reward function generate long-form stories with coherent plot arcs and unexpected, satisfying twists that outperform standard autoregressive generation across both automatic and human evaluation metrics?",
        "motivations": "Long-form story generation demands maintaining narrative coherence across many paragraphs while simultaneously introducing compelling, satisfying plot twists. Standard autoregressive large language models are fundamentally myopic, planning only one token or sentence at a time, which leads to boring, predictable trajectories or, when forced toward randomness, incoherent plots that lose track of earlier context. Search algorithms such as Monte Carlo Tree Search have revolutionized planning in games and structured environments, but their application to open-ended language generation has been limited by the absence of reliable narrative reward signals. Existing story generation methods either rely on greedy decoding that produces flat narratives or employ hierarchical approaches that still lack explicit long-horizon planning. A search algorithm guided by a purpose-built narrative reward could explicitly balance coherence and surprise over long horizons, enabling stories that are both logically consistent and dramatically engaging. This gap between game-theoretic planning and creative language generation remains largely unaddressed.",
        "methods": "This work would develop StoryPlan, a framework integrating Monte Carlo Tree Search with a large language model for long-horizon narrative planning. The LLM would serve as the policy network, proposing candidate next plot points, defined as one-to-two-sentence summaries, rather than individual tokens. The MCTS would expand a tree of possible future plot points over multiple search iterations. A separate value network, implemented as a fine-tuned LLM, would evaluate each node along two complementary dimensions: Coherence, measuring whether a given plot point logically follows from the accumulated narrative context, and Surprise, quantifying how much the plot point deviates from the expected trajectory implied by prior events. The Upper Confidence Bound selection formula would be modified to explicitly trade off exploitation of coherent, high-probability paths with exploration of surprising, lower-probability alternatives. A temperature-controlled weighting parameter would allow dynamic adjustment of this trade-off during search. Once the MCTS identifies an optimal path from the premise to a satisfying story ending, the LLM would perform surface realization, expanding the chosen sequence of plot points into final polished prose. This decoupling of planning from realization would allow the search to operate efficiently in a compact semantic space.",
        "experiment_plan": "The evaluation would focus on the story ending generation task using the CMU Movie Summary Corpus, where the goal is to generate a satisfying ending given a movie premise. Baselines would include standard LLM generation with GPT-4 and Llama-3-70B under both greedy and nucleus sampling, as well as hierarchical generation models that decompose stories into outline-then-prose stages. Automatic metrics would include story-cloze test accuracy for narrative logic assessment and a custom Twist-Plausibility score rated by a GPT-4 judge that evaluates whether generated twists are both surprising and consistent with the premise. Human annotators would evaluate 200 stories in pairwise comparisons against all baselines along three dimensions: Coherence, Plot Twist Surprise, and Overall Enjoyment. Ablation studies would isolate the contribution of search depth, the relative weighting of Coherence versus Surprise in the modified UCB formula, and the choice of plot point granularity, ranging from single sentences to multi-sentence paragraph summaries. Additional analysis would examine failure cases where excessive surprise degrades coherence."
      }
    },
    {
      "final_decision": "standard_reject",
      "source": "llm",
      "llm_name": "Qwen3.5-35B-A3B",
      "idea_card": {
        "title": "The Human-AI Creative Disentanglement Benchmark for Authorship Attribution and Stylistic Analysis",
        "research_question": "To what extent can current generative language models be reliably distinguished from human writers using a comprehensive suite of stylistic, semantic, and syntactic features across diverse creative writing genres?",
        "motivations": "Evaluating creativity often lacks ground truth, making it difficult to assess how AI differs from human creativity. Without knowing how AI differs from human creativity, we cannot improve alignment effectively. A benchmark quantifying these differences is crucial for responsible deployment and for understanding the limitations of synthetic creativity. Current evaluation metrics focus on fluency and coherence but fail to capture the nuanced stylistic markers that distinguish human authorship. This gap prevents researchers from identifying systematic biases in model generation and hinders the development of more authentic creative AI systems. Furthermore, the absence of standardized benchmarks makes it challenging to compare different models fairly or to track progress in human-like creative writing capabilities over time. Addressing this gap requires a rigorous framework that combines quantitative feature extraction with qualitative human evaluation to establish a reliable baseline for creative AI assessment.",
        "methods": "We curate a dataset of 1000 short stories labeled by human authorship with a 50/50 split, ensuring balanced representation across genres including science fiction, fantasy, and literary fiction. We extract features including syntactic complexity using dependency parsing, token diversity through type-token ratio calculations, and lexical sophistication via frequency-based vocabulary analysis. We train a multi-layer perceptron classifier to detect authorship and analyze feature importance using SHAP values to identify creative markers that models replicate or miss systematically. The method incorporates both traditional linguistic features and modern embedding-based representations from pre-trained models. We implement a two-stage training pipeline where initial feature extraction is followed by supervised classification with cross-validation. The classifier will be trained on 80% of the data with 20% reserved for testing, ensuring robust evaluation. We also develop a generative component that allows models to produce text for comparison against the human set, enabling direct evaluation of model outputs rather than just classification of existing text. This dual approach provides comprehensive insights into both detection capabilities and generation quality.",
        "experiment_plan": "Tasks involve classifying text source and generating text to compare against the human set. Metrics include classification accuracy and F1 score across different model families, with additional analysis of precision-recall trade-offs. Baselines include GPT-3.5, GPT-4, and Llama-3 variants, each tested at multiple temperature settings to evaluate robustness. We run experiments on the CreativityBench dataset we construct to measure generalization across genres, ensuring the benchmark captures diverse writing styles. We also perform ablation on model size to see if larger models mimic humans better, testing parameter counts from 7B to 70B. Error analysis will examine misclassification patterns to identify specific stylistic features that confuse classifiers. Robustness checks include testing on out-of-distribution samples and evaluating performance across different writing prompts. We will conduct human evaluation studies where participants rate the authenticity of generated text, providing qualitative validation of our quantitative findings. Additionally, we will analyze feature importance scores to understand which linguistic characteristics most strongly predict human authorship. This comprehensive evaluation framework will provide actionable insights for improving creative AI systems and establishing reliable benchmarks for future research."
      }
    }
  ]
}
\end{tcblisting}

\section{LLM-Based Idea Extraction and Anonymization Pipeline}
\label{sec:appendix_llm_idea_extraction}

We convert each source paper into a structured idea card before training or evaluation.
Each idea card contains a title, research question, motivations, methods, and an experiment plan, following the template in Figure~\ref{fig:idea_template}.
This format presents the idea as a minimal prospective proposal rather than a completed paper.
This prevents idea evaluators from receiving authorship information, experimental results, paper reviews, or other post-hoc evidence that could confound their judgment of the idea itself.

The pipeline comprises two stages. 
First, GLM-5.2 reads parsed PDF text and extracts the full paper's content into a structured idea card.  
The extraction prompt explicitly prohibits the LLM from 
inventing methods, datasets, metrics, or baselines that are not presented by the paper, and we apply a retry budget when the output JSON schema is broken or incomplete. 
Then, GLM-5.2 rewrites the idea
card into an anonymous idea proposal. It replaces the paper's newly proposed method, model, module, and benchmark names with anonymized technical descriptions.

To verify anonymization, we apply both LLM and deterministic safeguards.  
Along with the anonymized idea, GLM-5.2 also returns an anonymization map that records every paper-specific short name it replaces.
A case-insensitive term-matching checker then verifies that none of these original terms remain in the anonymized card.
Additional validators reject title n-gram reuse, name-like acronyms, completed-paper language such as reported results,
and scope drift relative to the faithful card.
A semantic-fidelity checker further compares the research object, intervention, and evaluation setting between the
faithful and anonymized versions.
Failed checks trigger a repair round, and we retain the original card, anonymized card, anonymization map, and
verification record as an audit trail.
All evaluated baselines of \IdeaRewardBench receive only the final anonymized card as input.
For retrieval-enabled evaluators, we further restrict retrieved evidence by setting the search cutoff to the day before the earliest initial release date among all ideas in the evaluated instance.

All prompts used in the LLM-based idea extraction and anonymization pipeline are provided in Section~\ref{sec:appendix_prompts}.


\section{Benchmark Data Statistics of \IdeaRewardBench}
\label{sec:appendix_benchmark_data_statistics}

Table~\ref{tab:appendix_benchmark_stats} summarizes the final benchmark composition.
The review and impact-grounded splits contain both pair and quadruple instances, while the annotation-grounded split contains controlled pairwise comparisons.

\begin{table}[!htbp]
\centering
\small
\caption{Final \IdeaRewardBench data distribution.}
\label{tab:appendix_benchmark_stats}
\begin{tabular}{lrrrr}
\toprule
Split & Pairs & Quadruples & Instances & Ideas \\
\midrule
Review-grounded & 1,580 & 285 & 1,865 & 4,300 \\
Impact-grounded & 827 & 264 & 1,091 & 2,710 \\
Annotation-grounded & 100 & 0 & 100 & 200 \\
\midrule
Total & 2,507 & 549 & 3,056 & 7,210\\
\bottomrule
\end{tabular}
\end{table}

Table~\ref{tab:appendix_annotation_stats} further details the annotation-grounded split.
It contains an equal number of research-question-seeded and topic-seeded comparisons, and covers both human-vs-LLM and LLM-vs-LLM settings.

\begin{table}[!htbp]
\centering
\small
\caption{Annotation-grounded split composition.}
\label{tab:appendix_annotation_stats}
\begin{tabular}{lrr}
\toprule
Subsplit & Pairs & Ideas \\
\midrule
Human-vs-LLM, research-question seeded & 25 & 50 \\
Human-vs-LLM, topic seeded & 25 & 50 \\
LLM-vs-LLM, research-question seeded & 25 & 50 \\
LLM-vs-LLM, topic seeded & 25 & 50 \\
\midrule
Total & 100 & 200 \\
\bottomrule
\end{tabular}
\end{table}

The 150 LLM-generated annotation-grounded ideas are generated by six LLMs: Qwen3.5-35B-A3B (39), Llama3.1-8B (30), GLM-5.2 (29), GPT-5.5 (28), DeepSeek-v4-pro (23), and Qwen3.8-27B (1).
The research-question-seeded and topic-seeded generation prompts are provided in Appendix~\ref{sec:appendix_idea_generation_prompts}.
The topic seeds used in this split are listed below.

\paragraph{Paper-extracted topic seeds.}
{\small
\begin{itemize}
\setlength\itemsep{0pt}
\setlength\parskip{0pt}
\item \textbf{Multi-objective alignment:} novel inference-time methods for large language models to balance competing human preferences.
\item \textbf{Long context:} novel methods to improve efficient reasoning and information recall over long contexts in large language models.
\item \textbf{Long-term memory:} scalable memory architectures for efficient long-context language modeling.
\item \textbf{Federated multimodal learning:} parameter-efficient methods to balance personalization and generalization in decentralized vision-language models.
\item \textbf{Dataset ownership:} novel methods to verify whether protected text data was used in language model pre-training.
\item \textbf{Multimodal reasoning:} resource-efficient methods to improve reasoning in vision-language models.
\item \textbf{Reinforcement learning preparation:} novel training strategies to prepare large language models for reasoning-oriented reinforcement learning.
\item \textbf{Code reasoning evaluation:} a novel benchmark for fine-grained semantic understanding over real-world programs.
\item \textbf{Machine unlearning:} efficient methods for reliable continual unlearning in large language models.
\item \textbf{Visual grounding:} novel inference-time methods to help vision-language models identify and use task-relevant visual evidence.
\item \textbf{Multi-step reasoning:} novel methods to improve multi-step reasoning in large language models.
\item \textbf{Backdoor defense:} robust methods to detect poisoned training samples under weak or stealthy backdoor attacks.
\item \textbf{Math reasoning:} novel data synthesis and post-training methods to improve large language model math problem solving.
\item \textbf{Software testing evaluation:} a novel benchmark for assessing generated test cases and their coverage of program faults.
\item \textbf{Honesty and calibration:} annotation-efficient methods to improve confidence calibration and honest uncertainty expression in large language models.
\item \textbf{Imperfect supervision:} novel methods to prevent undesirable behaviors during large language model fine-tuning with misspecified feedback.
\item \textbf{Scientific document reasoning:} a new benchmark for multimodal understanding in academic papers.
\item \textbf{Multimodal forecasting:} a novel, leakage-resistant benchmark for time-series prediction with textual context.
\item \textbf{Embedding privacy:} novel methods to protect sensitive concepts in text representations while preserving semantics and downstream utility.
\item \textbf{Vulnerability analysis:} novel datasets and methods for linking natural-language vulnerability reports to source-code patches.
\item \textbf{Fine-grained visual reasoning:} novel methods to improve vision-language model understanding of dense multi-panel content.
\item \textbf{Information-seeking agents:} process-level evaluation of evidence use and decision-making in large language model search agents.
\item \textbf{Model auditing:} novel methods to uncover knowledge that language models possess but do not explicitly disclose.
\item \textbf{Improving reasoning with RL:} novel methods to improve reinforcement learning for large language model reasoning.
\item \textbf{User simulation:} novel methods on realistic user simulators for evaluating tool-using agents under difficult and non-collaborative user interactions.
\end{itemize}
}

\paragraph{Expert-designed topic seeds.}
{\small
\begin{itemize}
\setlength\itemsep{0pt}
\setlength\parskip{0pt}
\item \textbf{LLM creative generation:} methods for evaluating or improving LLM performance on creative generation tasks.
\item \textbf{Benchmarking autonomous research agents:} methods for systematically evaluating autonomous agents on complex, long-horizon scientific research tasks.
\item \textbf{Balancing alignment and performance:} methods for better aligning LLMs with user preferences while minimizing degradation in task performance.
\item \textbf{Agent safety:} methods for evaluating or improving the safety of autonomous agents performing long-horizon tasks.
\item \textbf{Agent harness self-evolvement:} training or inference-time methods for agent harness self-evolvement.
\end{itemize}
}

\section{Training Data Statistics of \IdeaReward}
\label{sec:appendix_training_data_construction}

\paragraph{Review-based Pairs.}
The review-based subset contains 8,697 pairs derived from 2021-2025 ICLR, ICML 2025, and 2021-2024 NeurIPS submissions and OpenReview peer reviews.
We follow the same extraction, grouping, and accept-reject pairing pipeline as the review-grounded benchmark split.
For training, we only anonymize half of the pairs at random to preserve exposure to real method-naming conventions, and relax the required gap between the accepted and rejected submissions' average human-reviewer overall scores to \(0.0\).

\paragraph{Impact-based Pairs.}
The impact-based subset contains 5,487 pairs derived from historical arXiv papers released between January 2020 and January 2025.
We group papers by arXiv category, release time window, and semantic similarity following the same rules as the impact-grounded benchmark split, while anonymizing only half the ideas at random and relaxing the semantic similarity threshold to 0.93.
Since these training papers 
are already matched within three-month release intervals, we construct preferences directly using raw citations rather than normalized daily citations.

\paragraph{Controlled Synthetic Pairs.}
The controlled synthetic subset contains 5,486 pairs.
We first extract seed topics, research questions, and motivations from unused papers in the review-based subset.
Then, we generate comparable ideas from the same seed topic or the same seed research question and motivation, and assign preferences through diverse automated strategies: treating paper-derived human ideas as preferred over LLM-generated ideas, preferring ideas from stronger LLMs over weaker LLMs, using GPT-5.6-Sol as a pairwise preference judge, and rewriting both sides of unused gold human-human pairs with different LLMs while preserving their original preference labels.
To improve dimension awareness, we further construct degraded rejected ideas by selectively weakening novelty, feasibility, significance, or clarity.
We also perturb length and writing style for 20\% of randomly selected synthetic pairs to reduce reliance on superficial presentation cues.
The shared idea-generation prompts and dimension-degradation instructions are provided in Appendices~\ref{sec:appendix_idea_generation_prompts} and \ref{sec:appendix_dimension_degradation_prompt}.

Table~\ref{tab:appendix_training_stats} summarizes the final training set.
The 14,184 paper-derived pairs are nearly evenly split between original and anonymized idea-card renderings, with 7,088 original pairs and 7,096 anonymized pairs.

\begin{table}[!htbp]
\centering
\small
\caption{Final \IdeaReward training data distribution.}
\label{tab:appendix_training_stats}
\begin{tabular}{lrrl}
\toprule
Source & Pairs & Share & Preference signal \\
\midrule
Review-grounded & 8,697 & 44.2\% & Peer-review decisions and scores \\
Impact-grounded & 5,487 & 27.9\% & Citation-based impact comparisons \\
Controlled synthetic & 5,486 & 27.9\% & Automated controlled preferences \\
\midrule
Total & 19,670 & 100.0\% & -- \\
\bottomrule
\end{tabular}
\end{table}

\begin{table}[!htbp]
\centering
\small
\caption{Controlled synthetic training-pair construction.}
\label{tab:appendix_synthetic_training_stats}
\begin{tabular}{lr}
\toprule
Construction strategy & Pairs \\
\midrule
Human-vs-LLM default pairing & 826 \\
Human-vs-LLM with LLM-judge assignment & 865 \\
Dimension-specific degradation & 1,272 \\
Strong-vs-weak LLM tier pairing & 1,271 \\
LLM-vs-LLM with LLM-judge assignment & 871 \\
Label-preserving rewrite of gold pairs & 381 \\
\midrule
Total & 5,486 \\
\bottomrule
\end{tabular}
\end{table}

Within the dimension-specific degradation subset, the degraded dimensions are balanced across clarity (327 pairs), novelty (318), significance (317), and feasibility (310).

\section{Analysis of Superficial Shortcuts}
\label{sec:appendix_shortcut_analysis}

\paragraph{Surface-Heuristic Baselines.}
During benchmark development, we evaluate three simple heuristics based on idea-card length, dataset mentions, and baseline mentions.
The card-length heuristic favors cards with larger total word counts.
For the other two heuristics, we use GLM-5.2 to count the respective mentions and favor cards with larger counts.
Table~\ref{tab:appendix_shortcut_baselines} reports the resulting performance.

\begin{table*}[!htbp]
\centering
\small
\caption{Performance of simple surface-heuristic baselines.}
\label{tab:appendix_shortcut_baselines}
\resizebox{\textwidth}{!}{
\begin{tabular}{lccccc}
\toprule
Evaluator & Review pair acc. & Review quadruple PRA / top-1 & Impact pair acc. & Impact quadruple PRA / top-1 & Annotation pair acc. \\
\midrule
Card length & 50.1\% & 52.0\% / 27.2\% & 48.7\% & 47.4\% / 24.9\% & 55.0\% \\
Dataset mentions & 51.0\% & 56.7\% / 29.0\% & 54.0\% & 53.8\% / 35.0\% & 63.0\% \\
Baseline mentions & 57.0\% & 53.8\% / 33.8\% & 44.0\% & 41.0\% / 13.0\% & 69.0\% \\
\bottomrule
\end{tabular}
}
\end{table*}

Across the review-grounded and impact-grounded splits, all three heuristics remain close to random performance.
Although dataset and baseline mentions are somewhat predictive on the annotation-grounded split, their performance remains substantially below that of \IdeaReward.
Thus, these simple surface cues do not account for \IdeaReward's performance.

\paragraph{Authorship-Reversal Challenge.}
To test whether \IdeaReward merely detects authorship and prefers human ideas, we construct 60 manually verified pairs that reverse the usual human-source preference.
We select weak human ideas from zero-citation papers or strong-reject submissions, generate LLM ideas for the same research questions, and manually verify that the selected LLM idea is stronger.
The gold-preferred idea is therefore always LLM-generated.

\begin{table}[!htbp]
\centering
\small
\caption{Accuracy on the authorship-reversal challenge.}
\label{tab:appendix_authorship_reversal}
\begin{tabular}{lr}
\toprule
Evaluator & Accuracy \\
\midrule
Always prefer human & 0.0\% \\
\IdeaReward-Llama & 71.6\% \\
\IdeaReward-Qwen & 76.7\% \\
\bottomrule
\end{tabular}
\end{table}

Both \IdeaReward models achieve over 70\% accuracy despite the reversed authorship preference, indicating that their predictions are not reducible to human-authorship detection.

\section{Data Licensing and Attribution}
\label{sec:appendix_data_licensing}

License screening forms part of our source-selection and data-construction pipeline.
For papers collected from OpenReview, we retrieve license information from the corresponding OpenReview submission record.
For papers collected from arXiv, we retrieve license information directly from arXiv.
We select source materials licensed for redistribution and adaptation during source selection.
For paper-derived data, we preserve per-source license and attribution metadata and follow the requirements of the applicable source licenses.

\section{Evaluation Details of \IdeaRewardBench}
\label{sec:appendix_evaluation_details}

For pointwise evaluators, we score each idea independently and induce all pair and quadruple rankings from the resulting scalar scores.
For pairwise-only evaluators, we query the evaluator directly on each pair; selecting the gold preferred idea receives 1, predicting a tie receives 0.5, and selecting the other idea receives 0.
For quadruples, we run all six round-robin comparisons, count a tie as half a win, and use each idea's Copeland win rate as its induced scalar score for top-1 accuracy and pairwise ranking accuracy.
This converts every evaluator into the same preference-order interface while preserving each method's native output format.

InnoEval~\citep{innoeval} is substantially more search-intensive than the other evaluated baselines.
Its pipeline performs external paper, web, and code search, followed by evidence report construction and grounded multi-perspective evaluation.
The released implementation depends on arXiv search, Serper/Google web search, Jina content extraction, Semantic Scholar, GitHub, and optional Kaggle credentials.
Because these APIs impose rate limits and make full-benchmark execution costly, we evaluate InnoEval on a fixed random subset: 50 review-grounded pairs, 50 review-grounded quadruples, 50 impact-grounded pairs, 50 impact-grounded quadruples, and all 100 annotation-grounded instances.
\IdeaReward and all other baselines are evaluated on the full 3,056-instance benchmark.
Detailed baseline-specific adaptations are described in Section~\ref{sec:appendix_baseline_adaptations}.

\begin{table*}[!htbp]
\centering
\small
\caption{InnoEval results on the fixed random subset. The review-grounded and impact-grounded evaluations each contain 50 pairs and 50 quadruples, and the annotation-grounded evaluation contains all 100 pairs.}
\label{tab:innoeval_subset_results}
\resizebox{\textwidth}{!}{
\begin{tabular}{lcccccccc}
\toprule
& \multicolumn{3}{c}{Review-grounded} & \multicolumn{3}{c}{Impact-grounded} & \multicolumn{2}{c}{Annotation-grounded} \\
\cmidrule(lr){2-4} \cmidrule(lr){5-7} \cmidrule(lr){8-9}
Model & Pair Acc. & Top-1 Acc. & PRA & Pair Acc. & Top-1 Acc. & PRA & Human-vs-LLM & LLM-vs-LLM \\
\midrule
InnoEval & 66.0 & 34.0 & 61.3 & 51.0 & 26.1 & 58.9 & 72.0 & 95.0 \\
\bottomrule
\end{tabular}
}
\end{table*}

\section{Baseline Adaptations}
\label{sec:appendix_baseline_adaptations}

All systems receive the same anonymous five-field idea card.  We preserve a
system's native scoring mechanism whenever it provides one. 
The specific adaptations for each baseline are outlined below.

\begin{itemize}
    \item \textbf{AI Scientist-v1.} We use its native idea schema and three
    ratings (interestingness, feasibility, and novelty), averaging them into a
    overall scalar.   Across all search workflows, returned evidence is filtered to the group's cutoff date.

    \item \textbf{AutoResearchClaw.} The original workflow reviews a
    paper and emits prose, so we render the card as a pseudo-paper and append the
    shared proposal-stage adaptation prompt in Appendix~K.4.  We preserve its
    three perspectives (methodology, domain, and statistics/rigor), then add a
    clearly labelled five-dimension score and independent overall scalar using
    the native review as rationale, solely to enable ranking and analysis.

    \item \textbf{ResearchAgent.} We call the native \textsc{ProblemValidator}
    rubric on the rendered research problem and rationale.  Its five native
    1--5 ratings (clarity, relevance, originality, feasibility, and
    significance) are averaged as an overall score; no references or entities are supplied.

    \item \textbf{InternAgent.} We run the upstream assessment-only
    \textsc{RankingAgent} once on each anonymous group with its default
    criterion weights.  The SurveyAgent is excluded because it has no
    time-filtered retrieval policy.

    \item \textbf{CycleReviewer and DeepReviewer.} These paper-review systems
    receive the same pseudo-paper rendering and shared proposal-stage adaptation
    prompt.  

    \item \textbf{SciJudge.} We render cards into the model's native
    title--abstract--date format and apply the shared proposal-stage
    adaptation prompt.  The title is the anonymized card title; the abstract deterministically
    combines the card's research question, motivation, method, and evaluation
    plan; and the date is fixed to 2026-02-16 and shared by all candidates in a run. 

    \item \textbf{EvoScientist-skills.} We run EvoScientist with the native \texttt{research-ideation} skill and invoke only its Step-5 pairwise 
  four-dimension judgment and Elo aggregation, rather than the full idea-generation workflow.

    \item \textbf{Intern-Atlas.} We use its official HTTP interface on the
    anonymous card, retaining its returned overall and dimension scores.  A little amount of the card is capped only to satisfy the service's input limit.

    \item \textbf{ARIS.} We implement its workflow in Claude Code and preserve
    its novelty-check structur.  The original
    Gemini/Google components are replaced by the controlled
    GLM-5.2/Semantic-Scholar backend; the reported scalar is consequently an
    ARIS-derived novelty index.

    \item \textbf{InnoEval.} We invoke the official \texttt{SingleIdeaPipeline}
    on text input and parse its native evaluation result.   Search is filtered to the group's cutoff date; because of its high cost, it is run on the designated subset only. 
\end{itemize}

\section{Tie-rate Analysis}

\label{sec:appendix_tie_rate}
\begin{figure*}[!htbp]
    \centering
    \includegraphics[width=\textwidth]{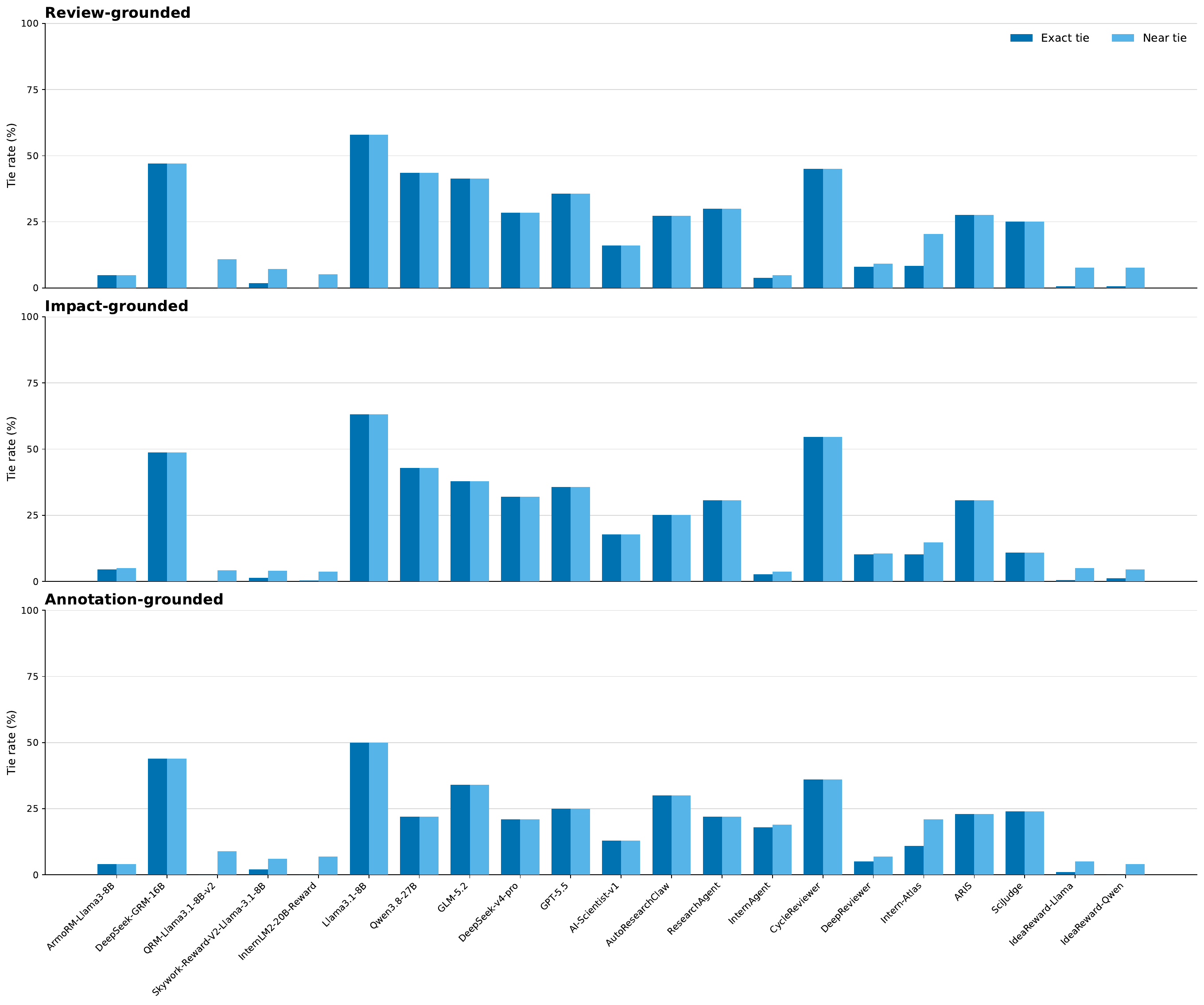}
    \caption{Exact (dark) and near (light) score-tie rates for the common 21
    scalar evaluators.  
    }
    \label{fig:appendix_tie_rate_three_tasks}
\end{figure*}

In this section, we investigate how often \IdeaReward and evaluated baselines predicts a tie between idea preferences of a same instance group. 
Figure~\ref{fig:appendix_tie_rate_three_tasks} compares score resolution for
the 21 pointwise evaluators evaluated on \IdeaRewardBench.  An exact tie means identical raw scores.  Near ties use a score difference below 0.1 after per-task normalization: within evaluator and
  conference for review-grounded pairs, within each impact quadruple for impact-grounded comparisons, and within evaluator over the valid
  annotation candidates.  Lower values are better.

The three panels show that tie-rate is consistent across all three benchmark splits. 
General RMs that emit continuous decimal scores tie far less than LLM-based methods constrained to integer ratings, whose exact-tie rates often exceed 40--60\%. 
Our IdeaReward models remain nearly tie-free under every construction: exact-tie rates of 0.63--0.70\% on review, 0.57--1.14\% on impact, and 0--1\% on annotation, with near-tie rates below 8\% throughout. 

\section{Impact-Grounded Score Distribution Analysis}
\label{sec:appendix_impact_sensitivity}

\begin{figure}[!hb]
\centering
\includegraphics[width=0.5\linewidth]{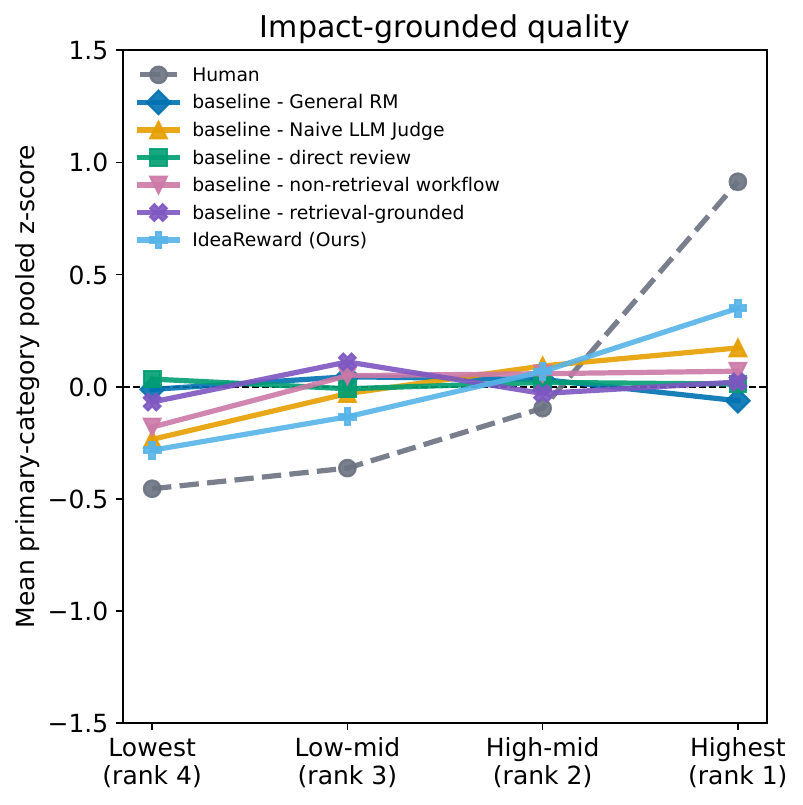}
\caption{Standardized score distributions across 264 impact-grounded
quadruples, ordered from lowest citation impact (rank~4) to highest
(rank~1).}
\label{fig:impact_sensitivity_appendix}
\end{figure}

This section presents the impact-grounded results of the score-distribution
analysis in Section~\ref{sec:score_distribution}. Scores are standardized
separately for each method and arXiv primary category. The figure uses the same
evaluator families and shared legend as the review-grounded panel
(Figure~\ref{fig:dimension_quality_combined}, right). In this panel, the dashed line labeled ``Human'' is computed from the raw citation counts of the source papers, standardized under the same arXiv-primary-category pooled z-score procedure.
We observe that \IdeaReward best aligns with human judgments across idea quality levels, while other baselines struggle to show a clear monotonic trend.

\section{Detailed Implementations of Dimensional Analysis}
\label{sec:appendix_dimensional_analysis}

We first conduct a meta-analysis of the official review criteria used by NeurIPS 2025, ICLR 2026, and ICML 2026.
The criteria can be mapped into four shared dimensions of idea quality: novelty, feasibility, significance, and clarity.
Table~\ref{tab:appendix_dimension_mapping} shows the venue-specific mapping.

\begin{table}[!htbp]
\centering
\small
\caption{Mapping venue-specific review criteria into unified idea-quality dimensions.}
\label{tab:appendix_dimension_mapping}
\begin{tabular}{llll}
\toprule
Unified dimension & NeurIPS 2025 & ICML 2026 & ICLR 2026 \\
\midrule
Novelty & Originality & Originality & GLM-5.2 completion \\
Feasibility & Quality & Soundness & Soundness \\
Significance & Significance & Significance & GLM-5.2 completion \\
Clarity & Clarity & Presentation & Presentation \\
\bottomrule
\end{tabular}
\end{table}

ICLR reviews provide soundness and presentation scores, but their contribution score conflates novelty and significance.
We therefore use a GLM-5.2-based judge to complete ICLR novelty and significance labels from the paper abstract, introduction, and reviewer contribution assessments; the prompt is shown in Section~\ref{sec:appendix_iclr_dimension_completion}.
A manual audit of 20 random ICLR completions found the completed labels consistent with the supplied review evidence.

Using these unified dimension scores, we construct 200 dimension-anchored preference pairs from the review-grounded split, with 50 pairs for each dimension.
Pairs are matched within the same venue and primary area.
For each dimension, 25 pairs have a chosen-minus-rejected average dimension-score gap of at least 2, and the other 25 have a gap in \([1,2)\), yielding both easy and medium-difficulty comparisons.

\section{Cost Analysis}

This section reports \IdeaReward's cost from two angles: building the
benchmark idea cards, and running the full evaluation suite across all
three splits.  Table~\ref{tab:construction_cost} estimates the one-time
cost of constructing the two main evaluation sets, counting one faithful
extraction, one proposal anonymization, and one scope-fidelity check per
selected paper at the GLM-5.2 list rates (CNY~8/M input, CNY~28/M
output)\footnote{GLM-5.2 API list pricing: \url{https://open.bigmodel.cn/pricing}.}.
Table~\ref{tab:cost_analysis} summarizes the provider-reported token
ledgers over the full evaluation suite: review-grounded (4,300 ideas),
impact-grounded (2,710 ideas), and annotation-grounded (200 ideas).  All
agentic evaluators (AI-Scientist-v1, AutoResearchClaw, ResearchAgent,
InternAgent, EvoScientist-skills, ARIS, InnoEval) use GLM-5.2 as their
backbone, so their cost follows the same GLM-5.2 list rates.
DeepSeek-v4-pro and GPT-5.5 use their respective commercial API list
prices, converted to CNY at 1~USD~=~7.1~CNY.  Locally served evaluators
(Llama3.1-8B, Qwen3.8-27B, CycleReviewer, DeepReviewer, SciJudge) run
on GPUs with no incremental API cost and are marked as local inference.
General reward models are excluded from this table.

\begin{table}[!htbp]
\centering
\small
\caption{Estimated GLM-5.2 cost of constructing the main evaluation cards, including faithful extraction, proposal anonymization, and scope-fidelity
  checking. Exceptional repair rounds are excluded.  }
\label{tab:construction_cost}
\begin{tabular}{lrrr}
\hline
Set & Cards & Estimated tokens (in / out) & Estimated cost (CNY) \\
\hline
Review-grounded & 4,300 & 69.57M / 12.19M & 897.88 \\
Impact-grounded & 2,710 & 42.52M / 7.88M & 560.80 \\
\hline
Total & 7,210 & 112.09M / 20.07M & 1,458.68 \\
\hline
\end{tabular}
\end{table}

\begin{table*}[!htbp]
\centering
\scriptsize
\caption{Provider-reported total tokens and estimated API cost. Each track
cell shows average tokens per idea / total tokens on that track; the
``All tracks'' column aggregates all three tracks. The final column
estimates the total API cost in CNY from provider list pricing; locally
served evaluators incur no API cash cost. InnoEval is run on a 200-group
review and impact subset plus all 100 annotation-grounded pairs.}
\label{tab:cost_analysis}
\resizebox{\textwidth}{!}{%
\begin{tabular}{lrrrrr}
\hline
Evaluator & Review & Impact & Annotation & All tracks & Est.\ cost (CNY) \\
\hline
Llama3.1-8B & 1,748 / 9,021,853 & 1,798 / 5,394,590 & 1,774 / 354,819 & 1,768 / 14,771,262 & local inference \\                       
Qwen3.8-27B & 3,547 / 18,302,434 & 3,520 / 10,561,217 & 3,517 / 703,357 & 3,537 / 29,567,008 & local inference \\
GLM-5.2 & 3,432 / 17,711,488 & 3,429 / 10,286,941 & 4,442 / 888,384 & 3,455 / 28,886,813 & 554.61 \\
DeepSeek-v4-pro & 4,069 / 20,997,867 & 3,252 / 9,755,033 & 4,176 / 835,131 & 3,778 / 31,588,031 & 490.40 \\
GPT-5.5 & 2,403 / 12,397,979 & 2,465 / 7,395,593 & 2,188 / 437,506 & 2,420 / 20,231,078 & 1,685.48 \\
AI-Scientist-v1 & 5,371 / 27,713,501 & 8,068 / 24,204,675 & 8,168 / 1,633,526 & 6,406 / 53,551,702 & 529.34 \\
AutoResearchClaw & 6,556 / 33,828,783 & 4,719 / 14,156,999 & 5,607 / 1,121,410 & 5,874 / 49,107,192 & 1,105.57 \\
ResearchAgent & 11,726 / 60,506,430 & 12,147 / 36,442,150 & 11,207 / 2,241,321 & 11,865 / 99,189,901 & 1,639.58 \\
InternAgent & 2,922 / 15,075,015 & 2,041 / 6,121,886 & 2,000 / 399,927 & 2,583 / 21,596,828 & 401.44 \\
CycleReviewer & 5,968 / 30,794,468 & 6,154 / 18,463,461 & 6,208 / 1,241,665 & 6,041 / 50,499,594 & local inference \\                            
DeepReviewer & 3,683 / 19,006,367 & 3,927 / 11,781,976 & 3,490 / 697,925 & 3,766 / 31,486,268 & local inference \\
SciJudge & 2,061 / 10,632,375 & 2,561 / 7,684,090 & 1,377 / 275,453 & 2,224 / 18,591,918 & local inference \\
EvoScientist-skills & 3,121 / 16,103,995 & 2,784 / 8,351,950 & 1,140 / 228,047 & 2,953 / 24,683,992 & 435.82 \\
ARIS & 5,556 / 28,667,784 & 5,375 / 16,124,059 & 12,715 / 2,542,935 & 5,662 / 47,334,778 & 590.52 \\
InnoEval & 219,704 / 65,911,181 & 244,518 / 73,355,273 & 126,370 / 12,637,012 & 217,005 / 151,903,466 & 1,949.44 \\
\hline
\end{tabular}
}
\end{table*}

\section{Human Annotation Information}
\label{sec:appendix_human_annotators}

All human annotation required in this work was jointly conducted by two
annotators, both of whom are AI researchers: Annotator A has more than three
years of research experience, and Annotator B has more than one year of research
experience.  For each annotation task, the two annotators first annotated
independently, then cross-checked their results together and resolved conflicts through discussion. In practice, their conflict rate was consistently below
20\%.  
For the annotation-grounded preference labels, Cohen's $\kappa$ between the two annotators' independent annotations before disagreement resolution is 0.84.
All annotated data were filtered to match the annotators' areas of
expertise, ensuring that both annotators could make reliable judgments.

\section{Statistical Analysis of Best-of-N Idea Selection}
\label{sec:appendix_bon_significance}

We use two-sided exact binomial sign tests over wins and losses after excluding ties.
We also report bootstrap 95\% confidence intervals for the non-tie win rates.

\begin{table}[!htbp]
\centering
\small
\caption{Statistical analysis of best-of-N idea selection.}
\label{tab:appendix_bon_significance}
\begin{tabular}{lrrrrrl}
\toprule
Baseline & Win & Loss & Tie & Win rate & $p$ & 95\% CI \\
\midrule
InternLM2-20B-Reward & 31 & 7 & 22 & 0.816 & 0.0001 & $[0.684,0.921]$ \\
GPT-5.5 & 19 & 11 & 30 & 0.633 & 0.20 & $[0.467,0.800]$ \\
GLM-5.2 & 22 & 10 & 28 & 0.688 & 0.050 & $[0.531,0.844]$ \\
\bottomrule
\end{tabular}
\end{table}

\section{Prompts}
\label{sec:appendix_prompts}


\subsection{Faithful Idea Extraction Prompt}

The extraction stage uses the following system instruction and user template.
It creates a structured, source-faithful idea card before any anonymization is
attempted.

\begin{tcblisting}{idearewardprompt}
SYSTEM
You are a meticulous research analyst. You read a machine-extracted paper text and distill it into a strict JSON structure. Be faithful to the paper: do not invent methods, datasets, metrics, or baselines that are not stated. Use the paper's own terminology in this pass (do NOT anonymize yet). Describe the proposed method at proposal level: the core claim, mechanism, and main design steps, not full implementation details. Reply with ONLY a JSON object, no prose, no markdown fences.

USER
Extract the following JSON from the paper text below.

Schema (fill every field; use "" if truly absent):
{
  "research_question": "the research gap / limitation this idea addresses; do NOT include the paper's solution",
  "motivations": "why this problem matters; keep this short and avoid promotional hype",
  "methods": "the core method claim plus proposal-level mechanism and main design steps",
  "experiment_plan": "how to validate it: tasks/data/settings, metrics, baselines, and key experiments"
}

Rules:
- "research_question" must be the gap only. It must NOT reveal the paper's own solution.
- "motivations" should explain significance in 1-3 sentences; do not add generic "why now" or expected-impact marketing.
- "methods" captures the idea itself: what is proposed and how it works. For a benchmark or dataset paper, include the data/task construction and evaluation protocol as the method, but keep it as a compact text field rather than a long component list.
- "experiment_plan" captures validation only: tasks/data/settings, metrics, baselines, key experiments, and necessary ablations. Do not include real experimental results.

PAPER METADATA:
title: [TITLE]
venue: [VENUE] [YEAR]

PAPER TEXT (machine-extracted, may contain artifacts):
[PAPER_TEXT]
\end{tcblisting}

\subsection{Idea Anonymization Prompt}

The second stage converts the faithful idea card into an anonymized idea:

\begin{tcblisting}{idearewardprompt}
SYSTEM
You edit research ideas for a blind benchmark. Convert a completed paper description into an anonymous, not-yet-executed research proposal. Remove paper-specific names, acronyms, method/module/system/model names, and newly introduced dataset or benchmark names. Preserve external, already-public benchmark, dataset, metric, and base-model names when they make the proposed evaluation scientifically concrete. Preserve technical meaning and enough detail for scientific judgment. Never invent technical content. Reply ONLY JSON.

USER
Rewrite this extracted idea as a blind research proposal.

Return exactly this JSON object:
{
  "title": "an informative generic proposal title",
  "research_question": "the unresolved research question",
  "motivations": "why the question matters",
  "methods": "what the proposed study would build or test and how it would work",
  "experiment_plan": "what the study would evaluate and compare",
  "anonymization_map": {"original paper-specific term": "generic replacement"}
}

Hard requirements:
- This is an IDEA PROPOSAL, not a paper summary. Use forward-looking language: "would investigate", "would develop", "could test", "the evaluation would".
- Do NOT write "the paper/work/study proposes", "we propose", "we introduce", "experiments show", "results demonstrate", reported scores, improvements, or completed outcomes.
- Do not retain any original title phrase verbatim. Do not retain a paper-specific acronym or coined method/module/system/model, or a dataset/benchmark introduced by the source paper.
- Retain external, pre-existing public benchmarks, datasets, metrics, and base models (for example MMLU, HumanEval, IFEval, TriviaQA, BLEU, GPT-4o) when they specify the experiment plan. These names are not identity leaks by themselves; do not place them in "anonymization_map".
- Preserve the original broad research scope. Generic task-domain terms (for example deep search agents, long-horizon information retrieval, adaptive retrieval) are not paper-specific names and may remain. Do not replace a broad research target with a narrower proxy task such as web navigation.
- The title must still name the same primary research object or capability as the faithful idea. It may be generic, but must not narrow, broaden, or shift the task setting.
- "anonymization_map" must list every removed paper-specific term. Apply every replacement consistently in all five visible fields. Replacements must be plain descriptive phrases; never create a new abbreviation or acronym.

ORIGINAL PAPER TITLE (private; never copy its phrases):
[SOURCE_TITLE]

EXTRACTED IDEA:
[FAITHFUL_IDEA_JSON]
\end{tcblisting}

\subsection{Naive LLM Judge Prompt}

For the Naive LLM Judge baseline, every backbone is given the same idea review rubric below.  
We form our idea-review prompt by drawing on the official reviewer guidelines of NeurIPS, ICLR, and ICML and adapting them to idea-level evaluation.
\texttt{[TOPIC]} and \texttt{[RENDERED\_IDEA\_CARD]} are replaced with the topic and anonymous idea card for the instance being judged.
The output is schema-validated, and final idea preferences are induced following rankings of the
\texttt{overall\_score}.

\begin{tcblisting}{idearewardprompt}
You are an expert academic reviewer evaluating an anonymized scientific research idea. Evaluate only the information provided, do not infer missing details, and do not attempt to identify the original paper or authors.

### Research Idea

**Topic:** [TOPIC]

**Research Idea Proposal:**
[RENDERED_IDEA_CARD]

### Evaluation Rubric

- **Novelty:** Does the research idea provide new insights, deepen understanding, or highlight important properties of existing methods? Is it clear how this work differs from previous contributions, with relevant citations provided? Does the work introduce novel tasks or methods that advance the field? Does the work offer a novel combination of existing techniques, and is the reasoning behind this combination well-articulated? Originality does not necessarily require an entirely new method: evaluating existing methods or improving efficiency or fairness can also be valuable.

- **Feasibility:** Is the proposed method technically sound and realistically implementable? Are the methods appropriate? Are the experiments well-designed, and can they adequately test the central claims?

- **Significance:** Does this research idea address an important problem? Does it advance scientific understanding, capabilities, or practice? Could it influence future research or applications? Is the scope of impact broad or specialized, and is that appropriate for the contribution?

- **Clarity:** Are the research question, motivation, method, and experimental plan clearly written? Is sufficient information provided to understand and evaluate the idea?

- **Overall:** Independently assess the overall quality of the research idea and its potential to yield an acceptance-worthy research paper if executed as proposed, based on its strengths and weaknesses.

### Evaluation Procedure

First, provide a thorough assessment of strengths and weaknesses, touching on novelty, feasibility, significance, and clarity. Then assign all scores based on that assessment.

### Score Scale

Dimension scores: 4 = Excellent; 3 = Good; 2 = Fair; 1 = Poor.

Overall score:
10 = Exceptional strong accept; worthy of special recognition.
9 = Strong accept; among the strongest accepted ideas.
8 = Clear accept; strong contribution suitable for acceptance.
7 = Accept; merits clearly outweigh weaknesses.
6 = Marginally above the acceptance threshold.
5 = Borderline; merits and weaknesses are approximately balanced.
4 = Marginally below the acceptance threshold.
3 = Reject; substantive weaknesses outweigh merits.
2 = Clear reject; major flaws substantially undermine the idea.
1 = Strong reject; fundamentally flawed or insufficiently specified.

### Output Format

Return only one valid JSON object. Keep strengths and weaknesses concise and specific. Use integer scores and do not include additional fields.

{
  "strengths": ["Review of concise substantive strengths"],
  "weaknesses": ["Review of concise substantive weaknesses"],
  "scores": {
    "novelty": "integer from 1 to 4",
    "feasibility": "integer from 1 to 4",
    "significance": "integer from 1 to 4",
    "clarity": "integer from 1 to 4"
  },
  "overall_score": "integer from 1 to 10"
}
\end{tcblisting}


\subsection{Paper-Reviewer Adaptation Prompt}

Paper-level reviewer baselines CycleReviewer and DeepReviewer receive the following additional instruction during inference.  It makes explicit that the submission is an unexecuted idea proposal,
so absent experimental results or a full-length manuscript should not be treated as deficiencies.

\begin{tcblisting}{idearewardprompt}
### Evaluation Scope

Please evaluate the submission as a research idea proposal rather than a completed paper. Do not require completed experimental results or a fully written manuscript. Instead, evaluate the novelty, feasibility, significance, and clarity of the proposed research direction, methodology, and experimental plan, based on its potential to yield an acceptance-worthy research paper if executed as proposed.
\end{tcblisting}

\subsection{ICLR Dimension-Completion Judge Prompt}
\label{sec:appendix_iclr_dimension_completion}

ICLR reviews do not natively separate dimensions of novelty and significance.  We therefore
use an LLM judge to complete the novelty and significance score labels. The LLM Judge
receives the abstract, introduction, and the human reviewers'
reviews as signals to judge the paper's novelty and significance.

\begin{tcblisting}{idearewardprompt}
SYSTEM
You are judging the novelty and significance of a machine-learning research contribution. Read the paper abstract and introduction alongside the ICLR reviewers' contribution assessments. Treat each reviewer's contribution score and written feedback as important evidence, then use the abstract and introduction to resolve what is novel versus what is significant. Score each dimension independently on the ICLR-style 1--4 scale, where 1 is clearly weak, 2 is limited, 3 is solid, and 4 is exceptional. Do not use paper title, authors, venue, decisions, or overall reviewer ratings. Return only JSON.

USER
Return exactly:
{
  "novelty": {"score": 1, "rationale": "brief", "evidence": ["verbatim span from the abstract/introduction"]},
  "significance": {"score": 1, "rationale": "brief", "evidence": ["verbatim span from the abstract/introduction"]}
}

ICLR reviewer contribution assessments (private labels/evidence):
[REVIEWER_CONTRIBUTION_EVIDENCE]

Paper abstract and introduction (do not use title, author, venue, or paper identity as a scoring cue):
[PAPER_ABSTRACT_AND_INTRODUCTION]
\end{tcblisting}

\subsection{LLM Idea Generation Prompts}
\label{sec:appendix_idea_generation_prompts}

For annotation-grounded data and controlled synthetic training data, we use two
idea-generation templates.  The research-question-and-motivation-seeded prompt
keeps the seed question and motivation fixed and asks the model to complete the
title, method, and experiment plan.  The topic-seeded prompt asks the model to
first formulate a research question and then generate the full five-field idea
card.  In the LLM-vs-LLM topic-seeded setting, the same instruction is applied
in a batch form that asks for exactly \texttt{[NUM\_IDEAS]} distinct idea cards
under a top-level \texttt{ideas} list.

\paragraph{Research-question-and-motivation-seeded generation.}
\begin{tcblisting}{idearewardprompt}
You are an expert researcher skilled at developing novel, rigorous, and publication-worthy research ideas.

Given a research question and its motivation as context, propose a novel and plausible research idea that directly addresses the stated problem.
The proposal should have the potential, if successfully executed, to support a strong publication at a top-tier AI/ML conference such as ICLR, ICML, or NeurIPS.

Your final response should contain exactly one valid JSON object and no other text. The JSON object must:
- Be parseable by `json.loads`.
- Use double quotes for every key and string value.
- Contain exactly these three top-level keys: "title", "methods", "experiment_plan".
- Use a single string value for every key; do not use arrays, nested objects, markdown, comments, or trailing commas.

Required JSON shape:
{
  "title": "a concise research-paper-style title",
  "methods": "a detailed description of the proposed approach",
  "experiment_plan": "a concrete plan for evaluating the proposed approach, including concrete tasks or datasets, metrics, baselines, ablations, and other analysis experiments"
}

Requirements:
- Use forward-looking proposal language such as "would investigate", "would develop", and "could test".
- Your proposal should only describe the proposed approach and evaluation plan. Do NOT report scores, improvements, or completed outcomes.
- Keep the broad research area and motivation aligned with the context.
- Make methods and experiment_plan detailed enough for scientific judging.

The example below demonstrates the expected level of specificity and output format.

Example research question:
Can a single transformer architecture be pretrained on diverse relational databases with heterogeneous schemas and directly applied to unseen datasets and tasks in a zero-shot manner, without task- or dataset-specific fine-tuning or retrieval of in-context examples?

Example motivation:
Relational databases are the dominant repository of structured enterprise information, yet they lack foundation models that transfer across datasets and tasks. Prior relational deep learning approaches using GNNs or transformers remain tightly coupled to specific schemas and fail to generalize to new databases. Existing tabular foundation models generalize across datasets but cannot capture rich relational structure such as primary-foreign key links. Serializing databases into text for LLMs suffers from scalability issues and distributional mismatch with LLM pretraining data.

Example response:
{
  "title": "Relational Transformer: Toward Zero-Shot Foundation Models for Relational Data",
  "methods": "The proposed study would represent each database cell as a token, with embeddings constructed from its value, column name, and table name. It would augment standard self-attention with column attention, feature attention, and neighbor attention over primary-foreign key links, and cast downstream tasks as masked token prediction to enable self-supervised pretraining.",
  "experiment_plan": "Experiments would use RelBench with leave-one-database-out transfer. Metrics would include AUROC for binary classification and R2 for regression. Baselines would include relational GNNs, relational transformers, text-serialized LLM baselines, and tabular foundation models. Ablations would isolate schema semantics, relational attention patterns, in-context task rows, and context length."
}

Now, please generate a new research idea proposal for the following context.

YOUR RESEARCH QUESTION:
[RESEARCH_QUESTION]

YOUR MOTIVATION:
[MOTIVATIONS]

[EXTRA_INSTRUCTION]
\end{tcblisting}

\paragraph{Topic-seeded generation.}
\begin{tcblisting}{idearewardprompt}
You are an expert researcher skilled at developing novel, rigorous, and publication-worthy research ideas.

Given a research topic as context, identify a timely and important research question closely related to the topic. Then, propose a novel and plausible research idea that directly addresses your formulated research question.
The proposal should have the potential, if successfully executed, to support a strong publication at a top-tier AI/ML conference such as ICLR, ICML, or NeurIPS.

Your final response should contain exactly one valid JSON object and no other text. The JSON object must:
- Be parseable by `json.loads`.
- Use double quotes for every key and string value.
- Contain exactly these five top-level keys: "title", "research_question", "motivations", "methods", "experiment_plan".
- Use a single string value for every key; do not use arrays, nested objects, markdown, comments, or trailing commas.

Required JSON shape:
{
  "title": "a concise research-paper-style title",
  "research_question": "a concrete research question that is closely related to the topic and motivates the proposed idea",
  "motivations": "a detailed explanation of why the research question matters and what gap it addresses",
  "methods": "a detailed description of the proposed approach",
  "experiment_plan": "a concrete plan for evaluating the proposed approach, including concrete tasks or datasets, metrics, baselines, ablations, and other analysis experiments"
}

Requirements:
- Use forward-looking proposal language such as "would investigate", "would develop", and "could test".
- Your proposal should only describe the proposed approach and evaluation plan. Do NOT report scores, improvements, or completed outcomes.
- Keep the broad research area, technical problem, and intended contribution closely aligned with the topic.
- Make research_question, motivations, methods, and experiment_plan detailed enough for scientific judging.

The example below demonstrates the expected level of specificity and output format.

Example research topic:
Foundation models for relational databases: novel methods that transfer across heterogeneous schemas, databases, and prediction tasks.

Example response:
{
  "title": "Relational Transformer: Toward Zero-Shot Foundation Models for Relational Data",
  "research_question": "Can a single transformer architecture be pretrained on diverse relational databases with heterogeneous schemas and directly applied to unseen datasets and tasks in a zero-shot manner, without task- or dataset-specific fine-tuning or retrieval of in-context examples?",
  "motivations": "Relational databases are the dominant repository of structured enterprise information, yet they lack foundation models that transfer across datasets and tasks. Prior relational deep learning approaches remain tightly coupled to specific schemas, while tabular foundation models cannot capture rich relational structure such as primary-foreign key links.",
  "methods": "The proposed study would represent each database cell as a token, with embeddings constructed from its value, column name, and table name. It would augment standard self-attention with column attention, feature attention, and neighbor attention over primary-foreign key links, and cast downstream tasks as masked token prediction to enable self-supervised pretraining.",
  "experiment_plan": "Experiments would use RelBench with leave-one-database-out transfer. Metrics would include AUROC for binary classification and R2 for regression. Baselines would include relational GNNs, relational transformers, text-serialized LLM baselines, and tabular foundation models. Ablations would isolate schema semantics, relational attention patterns, in-context task rows, and context length."
}

Now, please generate a new research idea proposal for the following context.

YOUR RESEARCH TOPIC:
[TOPIC]

[EXTRA_INSTRUCTION]
\end{tcblisting}

\subsection{Dimension-Specific Degradation Prompt}
\label{sec:appendix_dimension_degradation_prompt}

For dimension-specific synthetic training pairs, we append the following
controlled-degradation instruction to either generation prompts presented in
Appendix~\ref{sec:appendix_idea_generation_prompts}.  The placeholder
\texttt{[DEGRADE\_INSTRUCTION]} is filled with one of the four dimension-specific
instructions below.

\begin{tcblisting}{idearewardprompt}
Additional controlled-degradation requirement:
[DEGRADE_INSTRUCTION]

Important: The generated idea should still be a syntactically valid and plausible research proposal, but it should have a clear weakness in the specified dimension.
\end{tcblisting}

\begin{tcblisting}{idearewardprompt}
Novelty:
Make the idea intentionally weak in novelty: it should mostly recombine familiar methods without a clear new insight, contribution, or distinction from prior work, while still sounding like a plausible research proposal.

Feasibility:
Make the idea intentionally weak in feasibility: include an underspecified or technically fragile method, unrealistic assumptions, or an evaluation plan that would be difficult to execute reliably.

Significance:
Make the idea intentionally weak in significance: focus on a narrow or low-impact problem where the expected contribution would be incremental and unlikely to influence future research or practice.

Clarity:
Make the idea intentionally weak in clarity: keep it superficially coherent but make the research question, method, or evaluation plan vague, ambiguous, and insufficiently specified.
\end{tcblisting}

\end{document}

%% file: math_commands.tex
\usepackage{amsmath,amsfonts,bm}

\def\eqref#1{equation~\ref{#1}}

\def\1{\bm{1}}

\DeclareMathAlphabet{\mathsfit}{\encodingdefault}{\sfdefault}{m}{sl}
\SetMathAlphabet{\mathsfit}{bold}{\encodingdefault}{\sfdefault}{bx}{n}



%% file: tables/main_results.tex
\begin{table*}[t]
\centering
\definecolor{tagblue}{RGB}{0,150,200}
\definecolor{tagorange}{RGB}{245,120,20}
\definecolor{taggreen}{RGB}{45,150,75}
\newcommand{\tagP}{\textcolor{tagorange}{\texttt{\textbf{[P]}}}}
\newcommand{\tagR}{\textcolor{tagblue}{\texttt{\textbf{[R]}}}}
\newcommand{\tagT}{\textcolor{taggreen}{\texttt{\textbf{[T]}}}}
\newcommand{\best}[1]{\textbf{#1}}
\newcommand{\second}[1]{\underline{#1}}
\caption{Main results on \IdeaRewardBench. \tagP{} denotes methods that use pairwise evaluation, \tagR{} denotes methods that support external-knowledge retrieval, and \tagT{} denotes methods trained on research idea-related data.}
\label{tab:main_results}
\resizebox{\textwidth}{!}{
\begin{tabular}{lcccccccc}
    \toprule
    & \multicolumn{3}{c}{Review-grounded} & \multicolumn{3}{c}{Impact-grounded} & \multicolumn{2}{c}{Annotation-grounded} \\
    \cmidrule(lr){2-4} \cmidrule(lr){5-7} \cmidrule(lr){8-9}
    Models & Pair Acc. & Top-1 Acc. & PRA & Pair Acc. & Top-1 Acc. & PRA & Human-vs-LLM & LLM-vs-LLM \\
    \midrule
    \multicolumn{9}{l}{\textit{General RMs}} \\
    \midrule
    ArmoRM-Llama3-8B & 48.4 & 19.9 & 46.1 & 48.9 & 22.9 & 47.4 & 16.0 & 44.0 \\
    QRM-Llama3.1-8B-v2 & 49.1 & 27.0 & 49.2 & 47.0 & 22.0 & 46.1 & 24.0 & 70.0 \\
    Skywork-Reward-V2-Llama-3.1-8B & 50.4 & 22.8 & 49.5 & 47.3 & 18.9 & 45.9 & 21.0 & 71.0 \\
    InternLM2-20B-Reward & 51.5 & 24.9 & 49.6 & 50.6 & 24.2 & 50.7 & 24.0 & 62.0 \\
    DeepSeek-GRM-16B & 49.0 & 24.6 & 48.3 & 50.8 & 23.5 & 50.0 & 41.0 & 43.0 \\
    PairRM\tagP & 50.3 & 23.2 & 49.7 & 43.7 & 20.1 & 45.5 & 12.0 & 78.0 \\
    pair-preference-model-LLaMA3-8B\tagP & 49.2 & 16.9 & 46.1 & 45.9 & 22.5 & 47.7 & 8.0 & 70.0 \\
    RRM-7B\tagP & 51.9 & 22.6 & 49.9 & 46.1 & 22.7 & 47.2 & 34.0 & 64.0 \\
    \midrule
    \multicolumn{9}{l}{\textit{Naive LLM Judge}} \\
    \midrule
    Llama3.1-8B-Instruct & 49.3 & 27.5 & 51.8 & 50.3 & 24.5 & 49.7 & 45.0 &53.0  \\
    Qwen3.8-27B & 62.8 & 32.1 & 60.5 & 57.6 & 30.9 & 56.4 & 81.0 & \second{91.0} \\
    GLM-5.2 & 60.6 & 35.1 & 59.4 & 58.1 & 32.9 & 56.2 & 68.0 & 78.0 \\
    DeepSeek-v4-pro & 58.4 & 24.9 & 55.1 & 56.9 & 33.6 & 56.2 & 67.0 & 76.0 \\
    GPT-5.5 & 58.0 & 29.2 & 56.2 & 58.8 & 34.1 & 56.4 & 79.0 & 82.0 \\
    \midrule
    \multicolumn{9}{l}{\textit{Specialized Idea Evaluators}} \\
    \midrule
    AI Scientist\tagR & 56.5 & 24.6 & 53.5 & 52.2 & 25.1 & 51.1 & 78.0 & 77.0 \\
    AutoResearchClaw & 57.8 & 31.3 & 55.2 & 53.1 & 25.4 & 51.2 & 57.0 & 83.0 \\
    ResearchAgent & 54.5 & 30.2 & 54.5 & 52.5 & 23.3 & 51.9 & 73.0 & 85.0 \\
    InternAgent & 58.2 & 36.6 & 60.0 & 54.8 & 31.1 & 56.6 & 69.0 & 87.0 \\
    CycleReviewer\tagT & 51.7 & 27.5 & 53.2 & 50.1 & 27.6 & 48.9 & 52.0 & 60.0 \\
    DeepReviewer\tagT & 58.1 & 35.0 & 58.5 & 57.4 & 28.6 & 51.0 & 52.0 & 79.0 \\
    SciJudge\tagP\tagT & 51.6 & 24.5 & 48.7 & 60.6 & 31.8 & 54.1 & 20.0 & 52.0 \\
    Intern-Atlas\tagR & 51.8 & 26.3 & 51.1 & 46.4 & 25.4 & 51.8 & 57.0 & 50.0 \\
    EvoScientist-skills\tagP & 59.3 & 33.2 & 57.1 & 55.1 & 28.6 & 53.1 & 68.0 & 86.0 \\
    ARIS\tagR & 50.7 & 27.6 & 51.3 & 45.0 & 24.0 & 47.5 & 50.0 & 63.0 \\
    \midrule
    \multicolumn{9}{l}{\IdeaReward (ours)} \\
    \midrule
    \textbf{\IdeaReward-Llama\tagT } & \second{67.3} & \best{42.1} & \second{65.6} & \second{70.3} & \second{39.6} & \second{61.6} & \best{100.0} & \second{91.0} \\
    \textbf{\IdeaReward-Qwen\tagT } & \best{67.8} & \second{41.4} & \best{67.8} & \best{72.4} & \best{41.5} & \best{63.5} & \second{98.0} & \best{94.0} \\
    \bottomrule
\end{tabular}
}
\end{table*}